\documentclass[10pt,twocolumn,letterpaper]{article}

\usepackage[pagenumbers]{iccv} 
\definecolor{iccvblue}{rgb}{0.21,0.49,0.74}
\usepackage[pagebackref,breaklinks,colorlinks,allcolors=iccvblue]{hyperref}
\usepackage{float}
\usepackage{threeparttable}
\usepackage{multicol,multirow}
\usepackage{colortbl}  
\usepackage{tcolorbox}
\usepackage{amsmath}  
\usepackage{algorithm}
\usepackage{algorithmic} 
\usepackage{amssymb}  
\usepackage{setspace}
\usepackage{tikz}
\usepackage{setspace}
\usepackage{pifont}
\usepackage{placeins}
\usepackage{float, afterpage}
\usepackage{amsfonts}
\newcommand{\bx}{\mathbf{x}}
\DeclareMathOperator*{\argmax}{argmax}
\usepackage{booktabs}

\usepackage{setspace}

\def\confName{ICCV}
\def\confYear{2025}
\makeatletter

\renewcommand{\@makefntext}[1]{%
  \parindent 1em%
  \noindent%
  \hb@xt@1.8em{\hss\@makefnmark}#1%
}
\newcommand{\linebreakand}{%
  \end{@IEEEauthorhalign}
  \hfill\mbox{}\par
  \mbox{}\hfill\begin{@IEEEauthorhalign}
}

\makeatother

\title{When Small Guides Large: Cross-Model Co-Learning for Test-Time Adaptation}

\author{
\vspace{-10pt}
\begin{tabular}{@{}c@{\hspace{1.8em}}c@{\hspace{1.8em}}c@{}}
\parbox[t]{0.3\textwidth}{\centering
Chang'an Yi\textsuperscript{*}\\ 
Foshan University \\ 
Foshan, China \\ 
{\tt\small yi.changan@fosu.edu.cn}
} &
\parbox[t]{0.3\textwidth}{\centering
Xiaohui Deng\textsuperscript{*}\\ 
Foshan University \\ 
Foshan, China \\ 
{\tt\small 2112353032@stu.fosu.edu.cn}
} &
\parbox[t]{0.3\textwidth}{\centering
Guohao Chen\textsuperscript{*}\\ 
\makebox[\linewidth]{Nanyang Technological University} \\ 
Singapore \\ 
{\tt\small  guohao.chen@ntu.edu.sg}
} \\[10ex] 
\parbox[t]{0.3\textwidth}{\centering
Yan Zhou\\ 
Foshan University \\ 
 Foshan, China\\ 
{\tt\small zhouyan791266@fosu.edu.cn}
} &
\parbox[t]{0.3\textwidth}{\centering
  Qinghua Lu\\ 
Foshan University \\ 
Foshan, China \\ 
{\tt\small qhlu@fosu.edu.cn}
} &
\parbox[t]{0.3\textwidth}{\centering
 Shuaicheng Niu\textsuperscript{\dag}\\ 
\makebox[\linewidth]{Nanyang Technological University} \\ 
 Singapore\\ 
{\tt\small shuaicheng.niu@ntu.edu.sg}
}
\end{tabular}
}

\begin{document}

\maketitle

\footnotetext[\numexpr\value{footnote}+1]{Equal contribution. \textsuperscript{\dag}Corresponding author.}
\vspace{-100pt} 

\begin{abstract}
Test-time Adaptation (TTA) adapts a given model to testing domain data with potential domain shifts through online unsupervised learning, yielding impressive performance. However, to date, existing TTA methods primarily focus on single-model adaptation. In this work, we investigate an intriguing question: how does cross-model knowledge influence the TTA process? Our findings reveal that, in TTA's unsupervised online setting, each model can provide complementary, confident knowledge to the others, even when there are substantial differences in model size. For instance, a smaller model like MobileViT (10.6M parameters) can effectively guide a larger model like ViT-Base (86.6M parameters). In light of this, we propose COCA, a \textbf{c}ross-m\textbf{o}del \textbf{c}o-le\textbf{a}rning framework for TTA, which mainly consists of two main strategies. 1) Co-adaptation adaptively integrates complementary knowledge from other models throughout the TTA process, reducing individual model biases. 2) Self-adaptation enhances each model’s unique strengths via unsupervised learning, enabling diverse adaptation to the target domain. Extensive experiments show that COCA, which can also serve as a plug-and-play module, significantly boosts existing SOTAs, on models with various sizes—including ResNets, ViTs, and Mobile-ViTs—via cross-model co-learned TTA. For example, with Mobile-ViT's guidance, COCA raises ViT-Base's average adaptation accuracy on ImageNet-C from 51.7\% to 64.5\%. The code will be publicly available.

\end{abstract}
   
\vspace{-0.25in}

\section{Introduction}
\label{sec:intro}


Deep neural networks have demonstrated remarkable performance when the training and test domains follow the same data distribution \cite{lu2022survey}. However, their accuracy drops significantly when testing data suffers from distribution shifts \cite{lu2022survey,liang2024comprehensive}.
To address this, numerous attempts have been explored, such as unsupervised domain adaptation \cite{ge2023unsupervised}, source-free domain adaptation~\cite{li2024comprehensive}, and domain generalization \cite{zhou2022domain}. However, in dynamic applications like autonomous driving, environmental conditions are constantly evolving, causing continuous data distribution shifts~\cite{li2023lwsis,sakaridis2021acdc}. To maintain stable and excellent performance, models need to adapt continuously to new conditions—an especially challenging task given the unpredictability of new environments~\cite{pan2020unsupervised}.  




\begin{figure*}[t]
\centering
\begin{subfigure}[t]{0.33\textwidth} 
    \centering
\includegraphics[width=\linewidth]{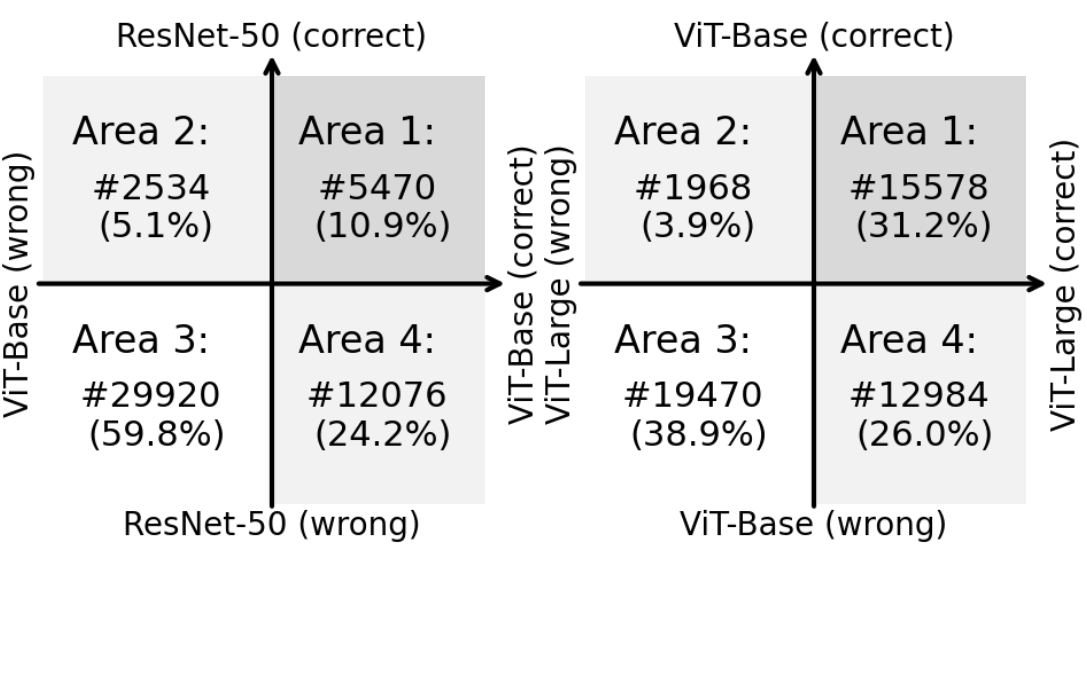}
\subcaption{Distinctness of model predictions
}
    \label{fig1a}
\end{subfigure}
\hfill
\begin{subfigure}[t]{0.34\textwidth}
    \centering
    \includegraphics[width=\linewidth]{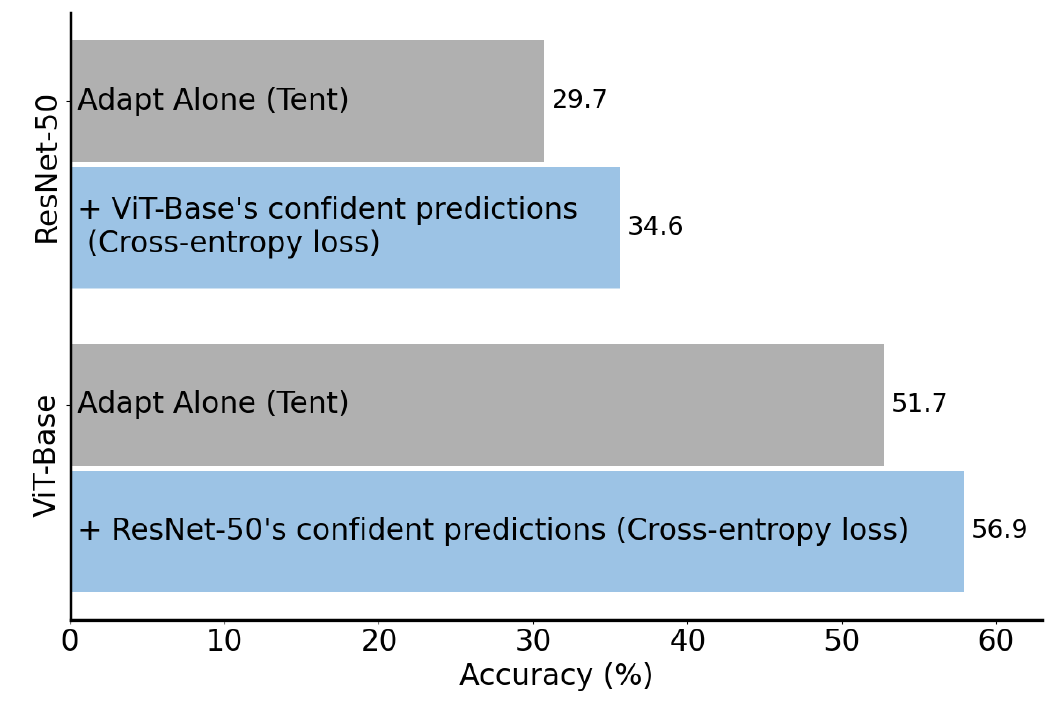}
    \subcaption{Benefits of complementary knowledge
}
    \label{fig1b}
\end{subfigure}
\hfill
\begin{subfigure}[t]{0.32\textwidth}
    \centering
    \includegraphics[width=\linewidth]{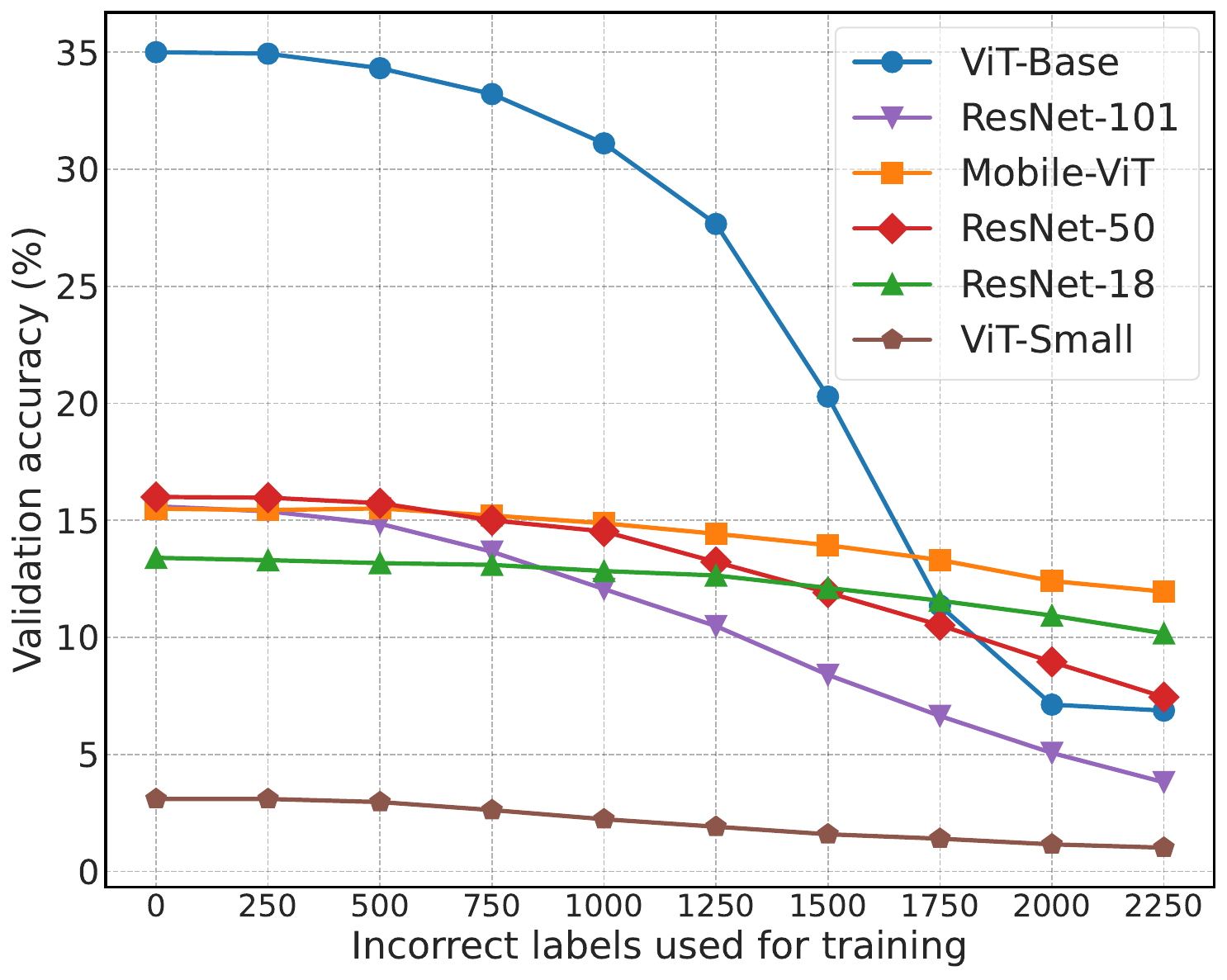}
    \subcaption{TTA robustness against noisy labels 
}
    \label{fig1c}
\end{subfigure}

\vspace{-0.1in}
\caption{Motivation for COCA. (a) Pre-trained models exhibit distinct strengths due to differences in training strategies, architectures, or sizes. (b) TTA~\cite{wang2020tent}, using complementary confident predictions from another model, significantly enhances adaptation performance. (c) Larger models are more vulnerable to erroneous guidance than smaller ones, highlighting differences in robustness during test time. Results are based on predicting ImageNet-C (Gaussian noise, level 5) with various ResNet-based~\cite{he2016deep} and Transformer-based~\cite{dosovitskiy2020image} models. Confident samples are filtered using an entropy threshold of 0.4*ln1000 following EATA~\cite{niu2022efficient}.}
\label{motifig1}
\vspace{-0.2in}
\end{figure*}

Recently, test-time adaptation (TTA) has emerged as a promising paradigm for tackling these challenges~\cite{sun2020test,liu2021ttt++,niu2023towards,lee2024entropy,liang2024comprehensive}. By directly leveraging test data through self-supervised or unsupervised objectives, TTA benefits from an online, source-free approach. This is especially true for fully TTA methods~\cite{wang2020tent,niu2022efficient}, which can be applied to any pre-trained model, thereby significantly broadening its appeal in real-world applications compared to earlier techniques. 
However, existing TTA methods~\cite{wang2020tent,niu2022efficient,lee2024entropy} typically focus on single-model adaptation and are excessively sensitive to pseudo label noise. 
Motivated by the success of co-learning~\cite{zhang2018deep} and knowledge distillation~\cite{hinton2015distilling,mirzadeh2020improved} in conventional supervised learning, we pose a natural question: Can co-learning also enhance TTA? 

To investigate this, we conduct two key experiments to examine the complementary capabilities of different models during TTA, revealing that: 1) \textit{Complementary knowledge between models}: As from Fig.~\ref{motifig1}(a), we show that different models can yield complementary predictions for the same data due to variations in training strategies, architectures, or sizes.
By leveraging the complementary confident samples from the other model to construct a pseudo cross-entropy loss during TTA, we significantly improve the adaptation performance on both the large and small model compared to Tent~\cite{wang2020tent}. 2) \textit{Differential robustness of models in TTA}: We evaluate the TTA robustness of various models by adapting them to test samples with randomly erroneous labels, using the same learning rate as in Tent~\cite{wang2020tent}. While larger models initially yield higher accuracy, they are more sensitive to incorrect pseudo labels during TTA, as shown in Fig.\ref{motifig1}(c). These findings indicate the potential of a small model to significantly improve the TTA performance and robustness of a larger model through cross-model co-learning.

In light of the above motivation, we propose a novel approach called COCA — \textbf{c}ross-m\textbf{o}del \textbf{c}o-le\textbf{a}rning for enhanced TTA, which mainly consists of two strategies, \ie, co-adaptation and self-adaptation. Co-adaptation integrates complementary knowledge between models during TTA to reduce individual biases. To achieve this, we introduce an adaptive temperature-scaled marginal entropy loss to promote mutual learning. The adaptive temperature is automatically determined via an anchor-guided alignment loss, using the larger model as an anchor. This enables robust knowledge aggregation, particularly when there is a significant size disparity between models, or the learning signals from the smaller models can otherwise be overstated or overlooked. Additionally, we propose a cross-model knowledge distillation loss, leveraging combined predictions as pseudo-labels to supervise both models and maximize the effective use of cross-model knowledge. On the other hand, self-adaptation allows each model to independently adapt using existing unsupervised learning objectives from single-model TTA, enhancing each model’s unique strengths and fostering diverse adaptation to the target domain. 

A key distinction of COCA from conventional methods is its bi-directional learning capability. Unlike typical teacher-student frameworks that distill knowledge unidirectionally from a powerful teacher to a weaker student~\cite{deng2019cluster,deng2023harmonious,hu2022teacher}, or co-learning schemes where models with similar architectures are involved~\cite{wang2022continual,dobler2023robust,zhou2023adaptive}, COCA enables every model—even those with weaker performance on out-of-distribution (OOD) data—to contribute valuable insights to others. This effective exchange of complementary knowledge allows COCA to perform well in various co-TTA scenarios, such as ResNet-18+ViT-Base and Mobile-ViT+ViT-Large, as shown in Table~\ref{allmodelssup} in Appendix. Our key contributions are as follows:

\begin{itemize}

     \item We reveal that two distinct models in TTA can mutually enhance each other. Each model extracts useful information from the other's predictions, improving TTA performance and stability, even when there is a large disparity in model size, such as between Mobile-ViT (10.6M parameters) and ViT-Large (304.2M parameters). 
     
     
     
     \item We propose COCA, a novel approach that seamlessly integrates co-adaptation and self-adaptation. Co-adaptation is achieved via adaptive temperature-scaled marginal entropy loss and cross-model distillation loss, enabling knowledge sharing to reduce biases. Self-adaptation enhances each model's unique strengths via unsupervised learning, enabling diverse adaptation to target data. 
     
     \item Extensive experiments show that COCA, which is also a plug-and-play module, significantly boosts TTA performance and stability across various co-adaptation scenarios. Moreover, it introduces minimal GPU overhead, and offers a flexible performance-efficiency tradeoff by incorporating smaller models like ResNet-18.

\end{itemize}

\vspace{-0.11in}
\section{Related Works}
\vspace{-0.03in}

\paragraph{Test-Time Adaptation} The goal of Test-Time Adaptation (TTA) is to leverage pre-trained models and adapt them to OOD data through fine-tuning, achieving robust performance in the target domain~\cite{liang2024comprehensive}. Recently, various methods have been proposed to tackle TTA. TEA~\cite{yuan2024tea} transforms the source model into an energy-based classifier to mitigate domain shifts. AdaContrast~\cite{chen2022contrastive} combines contrastive learning and pseudo-labeling to address TTA. DomainAdapter~\cite{zhang2023domain} adaptively merges training and test statistics within normalization layers. AdaNPC~\cite{zhang2023adanpc} is a parameter-free TTA approach based on a K-Nearest Neighbor (KNN) classifier, using a voting mechanism to assign labels based on $k$ nearest samples from memory. In contrast to traditional approaches, CTTA-VDP ~\cite{gan2023decorate} introduces a homeostasis-based prompt adaptation strategy, freezing the source model parameters during continual TTA. Additionally, FOA~\cite{niu2024test} devise a fitness function that measures test-training statistic discrepancy and model prediction entropy. 

Among the various TTA methods, entropy minimization has emerged as a prominent approach. For example, Tent~\cite{wang2020tent} minimizes the entropy of the model’s predictions on OOD data, adapting only the normalization layer parameters. Inspired by Tent, methods such as EATA~\cite{niu2022efficient}, MEMO~\cite{zhang2022memo}, SAR~\cite{niu2023towards}, DeYO~\cite{lee2024entropy}, and ROID~\cite{marsden2024universal2} have demonstrated strong performance in entropy-based TTA. However, most prior works leverage only the limited knowledge from a pre-trained model and thus result in constrained TTA efficacy, as shown in Table~\ref{mainres}.


\vspace{-15pt}
\paragraph{Co-Learning} Traditional co-learning paradigms include the teacher-student framework~\cite{hu2022teacher, beyer2022knowledge}, multi-modal settings~\cite{yin2023crossmatch}, and ensemble learning~\cite{yang2023survey}. In the teacher-student setting, knowledge is transferred unidirectionally from a stronger teacher to a weaker student. In contrast, COCA operates bidirectionally, enhancing overall performance by mutually improving both models. A key distinction between COCA and multi-modal settings is that in COCA, all models share the same input and task. In typical ensemble learning, models make independent predictions that are later aggregated. However, COCA enables direct inter-model influence throughout the entire adaptation process, fostering deeper collaboration. 

In the field of domain adaptation which is related to TTA, CMT~\cite{cao2023contrastive} proposes contrastive mean teacher to maximize beneficial learning signals. Harmonious Teacher~\cite{deng2023harmonious} is a sample reweighing strategy based on the consistency of classification. TTA methods like CoTTA~\cite{wang2022continual}, RoTTA~\cite{yuan2023robust}, and RMT~\cite{dobler2023robust} have successfully applied the teacher-student paradigm within the same model, yielding strong performance. In unsupervised domain adaptation, a closely related approach is AML~\cite{zhou2023adaptive}, which tackles this issue by adaptively switching the roles of the teacher and student, thereby alleviating the challenge of selecting a powerful teacher model. However, in TTA context, determining when to switch these roles becomes more complex, primarily due to the inconsistency in confidence levels between the outputs of different models. This inconsistency makes it difficult to establish a clear criterion for role-switching.

\vspace{-0.1in}
\section{Proposed Method}
\label{sec:Method Section}
\paragraph{Problem Statement} 
Let $\mathcal{D}^{train} = \{\left(\textbf{x}_i, \textbf{y}_i \right)\}_{i=1}^{N} \in \mathcal{P}^{train}$ denote the labeled training data from the source domain, where $\mathbf{x}$, $\mathbf{y}$ and $N$ represent the features, labels and data
amount, respectively. We consider two pre-trained source models, $f_{\theta_1}: \mathbf{x} \rightarrow \mathbf{y}$ and $f_{\theta_2}: \mathbf{x} \rightarrow \mathbf{y}$, parameterized by $\theta_1$ and $\theta_2$, respectively. These two models are well-trained on $\mathcal{D}^{\text{train}}$, and our goal is to adapt them to unlabeled, OOD target data in an online unsupervised manner. 

\vspace{-12pt}
\paragraph{Motivation} To fully exploit the potential of $f_{\theta_1}$ and $f_{\theta_2}$, a natural question arises: \textit{How do they benefit each other throughout the test-time adaptation process?} Unlike traditional paradigms such as the teacher-student framework~\cite{hu2022teacher}, ensemble learning~\cite{yang2023survey}, and multi-modal learning~\cite{yin2023crossmatch}, we aim to enable the models to mutually enhance each other in a bidirectional manner throughout the entire adaptation process. Additionally, the models share the same input and are designed for the same task. As shown in Fig.\ref{motifig1}, different models capture distinct facets of source knowledge due to variations in training strategies, architectures, or sizes—with larger models generally being more powerful. More importantly, Fig.\ref{motifig1} also reveals two key insights: 

\begin{itemize}

     \item \textbf{Different models offer complementary knowledge from training}.
     Though pre-trained models differ in sizes, they provide complementary confident knowledge to each other, which substantially enhances TTA performance. Taking ResNet-50 and ViT-Base as two peer collaborators, by harnessing the complementary knowledge from each other, the performance of ResNet-50 improves from 29.7\% to 34.6\%, while the performance of ViT-Base rises from 51.7\% to 56.9\% on ImageNet-C. 
     \item \textbf{Different models exhibit varying levels of robustness during TTA}. 
     While larger models are more accurate, smaller models can be more resistant to noisy learning signals, which commonly arise in complex TTA scenarios. 
     For example, when TTA is performed on 2,000 samples with incorrect pseudo labels, the performance of ViT-Base falls to a level even lower than that of ResNet-50.
\end{itemize}
These observations underscore the potential of exploring cooperative TTA mechanisms. However, achieving stable co-adaptation between models of different sizes remains challenging, due to the substantial disparity in their outputs.

\begin{figure}[t]
\centering
    \includegraphics[width=0.93\linewidth]{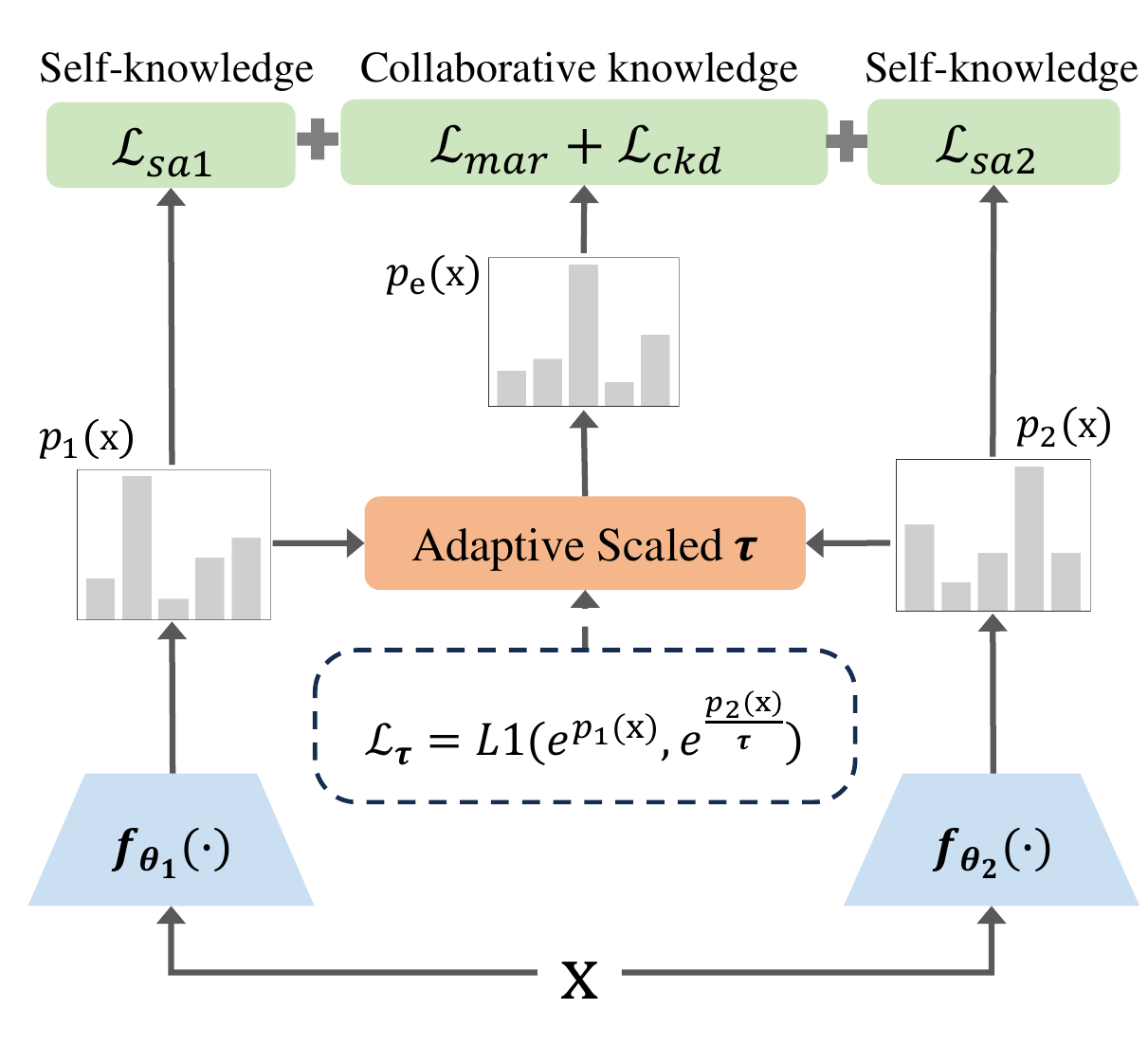}
    \caption{The overall COCA framework consists of two models that adapt by leveraging both their inherent self-knowledge and the adaptive collaborative knowledge shared between them. To facilitate a more effective and robust ensemble of their outputs—namely, $p_1(\bx)$ and $p_2(\bx)$—a learnable parameter, $\tau$, is introduced. In this example, we designate $f_{\theta_2}$ as the auxiliary model, and its output is divided by $\tau$ to enable co-learning.}
    \vspace{-0.15in}
\label{overview}
\end{figure}


\vspace{-11pt}\paragraph{Overall Framework} To leverage the unique strengths of different pre-trained models and address their varying robustness during test-time adaptation (TTA), we propose COCA, a cross-model co-learning approach that emphasizes an adaptive bidirectional cooperation mechanism. COCA employs two key strategies: co-adaptation and self-adaptation. The co-adaptation strategy harnesses the collaborative knowledge between models to mitigate individual biases during TTA, with adaptive alignment of outputs from models of different sizes further enhancing robustness. In contrast, the self-adaptation strategy refines each model’s unique strengths through unsupervised learning, enabling diverse and flexible adaptation to the target domain. An overview of COCA is shown in Fig.~\ref{overview}, and the corresponding pseudo-code is provided in Appendix~\ref{pscode}.

\begin{figure*}[t]
\centering
    \includegraphics[width=\linewidth]{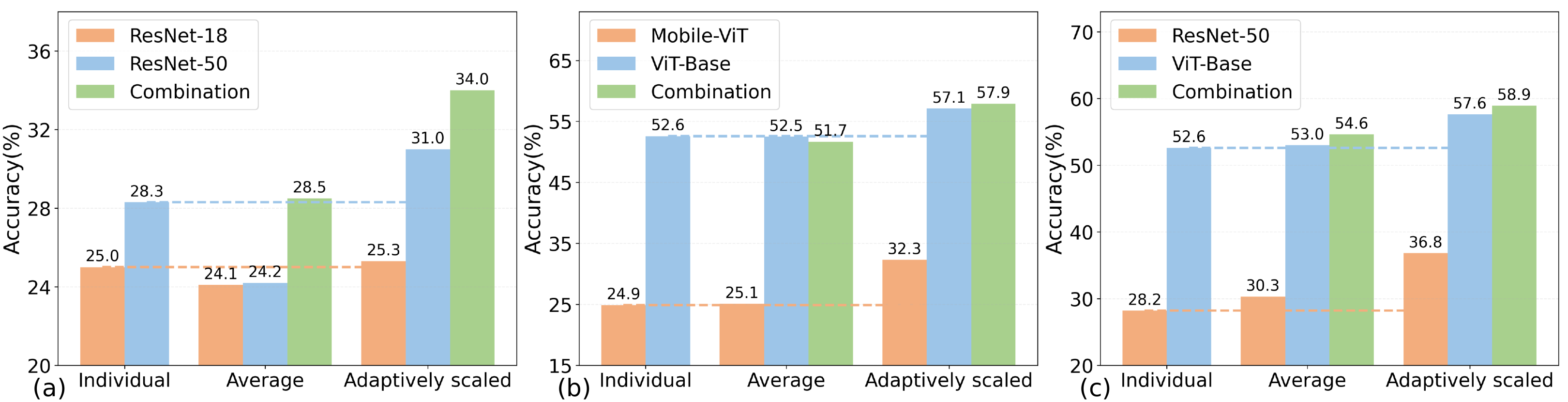}
    \vspace{-0.25in}
    \caption{The necessity of introducing $\tau$, a learnable parameter. All experiments are based on Tent~\cite{wang2020tent}. \textit{Individual} refers to adapting each model independently, as done in Tent. \textit{Average} involves combining the predictions of two models by averaging their output logits for marginal entropy minimization. Under this strategy, the performance improvement is limited. In contrast, \textit{Adaptively scaled} utilizes the parameter $\tau$ to adaptively combine the output logits, resulting in a substantial increase in overall performance. }
    \vspace{-0.1in}
\label{Tcompare}
\end{figure*}

\subsection{Cross-Model Collaborative Adaptation} 
\label{aoi}
\label{tintro}
Unlike traditional TTA methods that depend solely on the limited knowledge of a single pre-trained model, our approach focuses on enhancing TTA performance and stability through cross-model collaboration. This collaborative strategy is particularly vital in the context of online unsupervised adaptation, where individual models are prone to overfitting their inherent biases and accumulating errors, as demonstrated in Table~\ref{mainres}. 

However, co-learning between models during testing presents significant challenges, due to both 1) varying degrees of miscalibration~\cite{naeini2015obtaining, tomani2021post} and 2) notable performance gaps in model predictions when handling OOD data. Consequently, simply averaging the outputs of multiple models for TTA can degrade overall performance, as in Fig.~\ref{Tcompare}.

To make co-learning feasible at testing, we introduce a learnable diversity-aware scaling factor, $\tau$, which facilitates robust knowledge aggregation between large and small models for TTA. Specifically, in COCA, we identify the model with the larger parameter size as the \textit{anchor model}, and the other as the \textit{auxiliary model}, where larger models typically exhibit better in-distribution and out-of-distribution generalization~\cite{kim2023reliable}. 
During testing, we measure the discrepancies between the anchor predictions and the auxiliary predictions, which informs how trustworthy the auxiliary predictions are to determine the value of $\tau$. Formally, given anchor predictions logits $p_{a}(\bx)$ and auxiliary predictions logits $p_{s}(\bx)$, our goal is to learn a scaling factor $\tau$ in a unidirectional manner, so that:

\begin{equation}
    \arg \min_{\tau} \mathcal{L}_{s}(p_{a}(\bx), \frac{p_{s}(\bx)}{\tau}),
    \label{tloss}
\end{equation}
where $\mathcal{L}_{s}$ is a discrepancy function. Here, we adopt L1 loss to estimate this discrepancy, while projecting the prediction logits to the exponential space for discrepancy calculation, inspired by softmax. Then, $\mathcal{L}_{s}$ is formulated by:

\begin{equation}
    \mathcal{L}_{s}(p_a(\bx), p_s(\bx)) = \left |\left|e^{p_{a}(\bx)} - e^{p_{s}(\bx)} \right|\right|_1.
\end{equation}

We use this discrepancy loss to optimize \textit{only} the learnable factor $\tau$, and subsequently leverage $\tau$ for prediction aggregation, with the overall prediction $p_{e}(\bx)$ given by:
\begin{equation}
    p_{e}(\bx) = \frac{p_{e}'(\bx)}{T},~~\text{where}~p_{e}'(\bx)=p_{a}(\bx) + \frac{p_{s}(\bx)}{\tau}.
    \label{tau_ensemble}
\end{equation}
The learnable parameter $\tau$ serves to align the outputs of models with different sizes, thereby ensuring a more effective and robust adaptation process. The ensemble prediction is then utilized for marginal entropy minimization. Here, we introduce an adaptive balance factor $T$, defined as $\max p_{e}'(\bx) / \max p_{a}(\bx)$, to keep the max predictive logit unchanged after aggregation, maintaining a reasonable sharpness of $p_{e}(\bx)$.
Notably, the scaling factor $\tau$ does not modify the results of $p_{s}(\bx)$ but controls its contribution to the ensemble prediction $p_e(\bx)$ according to its reliability. \linebreak Experiments in Fig.~\ref{Tcompare} demonstrate the effectiveness of our $\tau$ to facilitate co-learning among models of various architectures and with different performance gaps.

Based on the ensemble prediction $p_{e}(\bx)$, we construct the cross-model co-adaptation objective of COCA as:
\begin{equation}
\label{colloss}
    \mathcal{L}_{col} = \mathcal{L}_{mar} + \mathcal{L}_{ckd},
\end{equation}
with $\mathcal{L}_{mar}$ to enhance the separability of data in the ensemble prediction $p_e(\bx)$  and $\mathcal{L}_{ckd}$ to improve each model through knowledge distillation. We depict them below.

\vspace{-5pt}\paragraph{Marginal Entropy Minimization} To improve the generalization of $p_e(\bx)$, our goal is to enhance the predictive confidence of $p_e(\bx)$, thereby learning a decision boundary that lies in the low-density sample region~\cite{grandvalet2004semi}. Following Tent~\cite{wang2020tent}, we adopt the entropy objective and apply it to $p_e(\bx)$ for marginal entropy minimization, defined as:
\begin{equation}
    \mathcal{L}_{mar} = - \sum_{c=1}^C p_{e}^c(\bx)\log p_{e}^c(\bx),
    \label{marginalent}
\end{equation}
where $C$ represents the number of categories.

\vspace{-10pt}\paragraph{Cross-Model Knowledge Distillation}
In addition to $\mathcal{L}_{mar}$ that enhances the overall prediction performance, we also introduce a knowledge distillation loss to provide more direct learning signals, transferring knowledge from $p_{e}(\bx)$ to each individual model. This maximizes the utilization of cross-model knowledge in $p_e(\bx)$, and improves the generalization of each model. Specifically, we derive pseudo-labels from $p_e(\bx)$ and compute a cross-model knowledge distillation loss for each model based on cross-entropy loss: 
\begin{align}
\label{ckdloss}
\mathcal{L}_{ckd} = \mathcal{L}_{CE}(p_a(\bx),\hat{y}) + \mathcal{L}_{CE}(p_s(\bx),\hat{y}),
\end{align}
where $\hat{y}= \argmax p_e(\bx)$ denotes the pseudo-label of the ensemble prediction $p_e(\bx)$.
\subsection{Self-Adaptation with Individual Knowledge}
\label{se}

The co-learning objective $\mathcal{L}_{col}$ aims to enhance TTA by reducing individual model biases. However, it enforces a uniform optimization direction across all models, which may overlook their unique capabilities. To address this, we further refine each model by leveraging its inherent knowledge, enabling us to harness their unique strengths and promote diverse adaptation to the target domain.

To this end, inspired by conventional single-model TTA methods, COCA adapts an individual model through the self/un-supervised learning objectives. Here, for simplicity, we adopt the entropy minimization loss from Tent~\cite{wang2020tent} and define the self-adaptation objective for each model as:
\begin{equation}
    \mathcal{L}_{sa} = - \sum_{c=1}^C (p_{a}^c(\bx)\log p_{a}^c(\bx) + p_{s}^c(\bx)\log p_{s}^c(\bx)).
    \label{selfloss}
\end{equation}
Note that $\mathcal{L}_{sa} $ is not limited to simple entropy minimization and COCA can seamlessly integrate with more advanced single-model TTA solutions, as verified in Table~\ref{mainres}.

In summary, COCA's overall optimization objective is the combination of the co-adaptation and self-adaptation objectives for all samples, which is formulated as:
\begin{equation}
\label{fullloss}
    \mathcal{L} =\mathcal{L}_{col} + \mathcal{L}_{sa}.
\end{equation}
We evaluate the effectiveness of each component in Table~\ref{ablation}, where each objective exhibits incremental improvement. We will discuss the influence of exploring the ratio between $\mathcal{L}_{col}$ and $\mathcal{L}_{sa}$ in Appendix~\ref{Ratio}.


\section{Experiments}
\label{sec:Experimental Section}

\begin{table*}[t]
    \setlength{\tabcolsep}{3pt} 
    \renewcommand{\arraystretch}{1.1} 
    \begin{center}
    \begin{threeparttable}
        \resizebox{\linewidth}{!}{
           \begin{tabular}{l|c|ccc|cccc|cccc|cccc|c}
\multicolumn{1}{c}{} & \multicolumn{1}{c}{} & \multicolumn{3}{c}{Noise} & \multicolumn{4}{c}{Blur} & \multicolumn{4}{c}{Weather} & \multicolumn{4}{c}{Digital} &  \\ \midrule
\multicolumn{1}{c|}{Methods} & Models & Gauss & Shot & Impul & Defoc & Glass & Motion & Zoom & Snow & Frost & Fog & Brit & Contr & Elastic & Pixel & JPEG & Avg. \\ \midrule
\multirow{2}{*}{Source Only} & ResNet-50 & 3.0 & 3.7 & 2.6 & 17.9 & 9.7 & 14.7 & 22.5 & 16.6 & 23.1 & 24.0 & 59.1 & 5.4 & 16.5 & 20.9 & 32.6 & 18.2 \\ 
 & ViT-Base & 35.0 & 32.1 & 35.8 & 31.4 & 25.3 & 39.4 & 31.5 & 24.4 & 30.1 & 54.7 & 64.4 & 48.9 & 34.2 & 53.1 & 56.4 & 39.7 \\ \midrule \midrule
\multirow{2}{*}{Tent~\cite{wang2020tent}} & ResNet-50 & 29.6 & 31.7 & 30.9 & 27.9 & 27.5 & 41.1 & 49.4 & 47.0 & 41.1 & 57.6 & 67.5 & 27.1 & 54.6 & 58.3 & 52.5 & 42.9 \\
 & ViT-Base & 51.7 & 51.5 & 52.9 & 51.9 & 47.9 & 55.7 & 49.3 & 10.2 & 18.5 & 67.3 & 73.0 & 66.4 & 52.0 & 64.7 & 63.2 & 51.7 \\ \midrule
\multirow{2}{*}{CoTTA~\cite{wang2022continual}} & ResNet-50 & 19.5 & 20.1 & 20.4 & 17.2 & 19.1 & 31.1 & 43.8 & 38.2 & 36.4 & 52.5 & 67.2 & 22.0 & 47.9 & 54.0 & 44.0 & 35.6 \\
 & ViT-Base & 56.8 & 56.3 & 58.8 & 45.7 & 49.8 & 61.1 & 49.7 & 57.6 & 58.8 & 69.7 & 75.1 & 60.8 & 59.2 & 69.2 & 67.3 & 59.7 \\ \midrule
\multirow{2}{*}{ROID~\cite{marsden2024universal2}} & ResNet-50 & 29.1 	&30.8 	&30.1 	&26.7 	&26.4 	&40.9 	&48.7 	&47.6 	&40.1 	&57.0 	&66.9 	&36.6 	&54.5 	&57.7 	&51.4 	&43.0 \\
 & ViT-Base & 52.0 	&51.7 	&52.8 	&47.4 	&48.4 	&56.9 	&52.3 	&56.2 	&53.4 	&68.2 	&73.8 	&65.1 	&57.0 	&67.1 	&64.3 	&57.8 \\ \midrule
 \multirow{2}{*}{DeYO~\cite{lee2024entropy}} & ResNet-50 & 35.5 & 37.4 & 36.9 & 33.5 & 32.9 & 46.8 & 52.5 & 51.6 & 45.8 & 60.0 & 68.6 & 42.4 & 58.0 & 60.9 & 55.5 & 47.8 \\
 & ViT-Base & 54.1 & 54.8 & 55.0 & 54.0 & 54.6 & 61.6 & 57.8 & 63.5 & 62.8 & 71.3 & 77.0 & 66.8 & 64.6 & 71.4 & 68.1 & 62.4 \\ \midrule
 
\textbf{} & ResNet-50 & 41.6 & 43.2 & 43.2 & 40.6 & 39.5 & 51.4 & 48.5 & 50.7 & 42.3 & 61.5 & 68.4 & 51.5 & 58.5 & 62.4 & 57.2 & 50.7 \\
\textbf{COCA (ours)} & ViT-Base* & 56.4 & 56.7 & 57.6 & 58.2 & 56.5 & 62.7 & 55.9 & 61.9 & 53.6 & 73.2 & 78.1 & 70.1 & 66.0 & 72.0 & 69.0 & 63.2 \\
 & \textbullet~Combined & \textbf{58.3} & \textbf{58.8} & \textbf{59.6} & \textbf{59.5} & \textbf{57.9} & \textbf{64.9} & \textbf{58.4} & \textbf{63.9} & \textbf{54.9} & \textbf{74.3} & \textbf{78.8} & \textbf{70.8} & \textbf{68.9} & \textbf{73.6} & \textbf{70.6} & \textbf{64.9} \\ \midrule \midrule
\multirow{2}{*}{EATA~\cite{niu2022efficient}} & ResNet-50 & 34.0 & 36.5 & 35.9 & 30.1 & 30.9 & 42.7 & 49.2 & 48.2 & 42.2 & 54.0 & 62.8 & 41.3 & 53.1 & 60.2 & 54.9 & 45.1 \\
 & ViT-Base & 55.6 & 56.2 & 56.7 & 54.1 & 53.9 & 58.7 & 58.5 & 62.5 & 60.7 & 69.6 & 75.7 & 68.8 & 63.2 & 69.5 & 66.6 & 62.0 \\ \midrule
 & ResNet-50 & 41.8 & 43.7 & 43.1 & 40.5 & 40.4 & 51.1 & 53.7 & 53.9 & 49.2 & 60.7 & 67.0 & 50.8 & 58.7 & 61.1 & 56.6 & 51.5 \\
EATA + \textbf{COCA (ours)} & ViT-Base* & 59.6 & 60.4 & 60.5 & 60.9 & 61.8 & 66.9 & 65.5 & 70.9 & 69.3 & 75.6 & 79.9 & 70.7 & 71.5 & 75.4 & 72.8 & 68.1 \\
 & \textbullet~Combined & \textbf{60.9} & \textbf{61.9} & \textbf{62.1} & \textbf{61.8} & \textbf{62.4} & \textbf{68.2} & \textbf{67.3} & \textbf{72.0} & \textbf{69.9} & \textbf{76.1} & \textbf{80.1} & \textbf{71.3} & \textbf{73.0} & \textbf{76.1} & \textbf{73.3} & \textbf{69.1} \\ \midrule \midrule
\multirow{2}{*}{SAR~\cite{niu2023towards}} & ResNet-50 & 34.0 & 35.9 & 35.1 & 30.9 & 30.0 & 46.5 & 51.7 & 50.9 & 44.9 & 59.6 & 67.9 & 41.3 & 57.3 & 60.3 & 55.0 & 46.8 \\
 & ViT-Base & 51.9 & 51.4 & 52.8 & 52.0 & 48.5 & 55.6 & 49.3 & 22.1 & 45.0 & 66.6 & 73.2 & 66.0 & 51.5 & 63.9 & 63.0 & 54.2 \\ \midrule
 & ResNet-50 & 40.9 & 42.9 & 42.3 & 39.9 & 39.2 & 51.2 & 52.6 & 52.4 & 45.6 & 61.7 & 68.6 & 51.2 & 58.8 & 62.3 & 57.4 & 51.1 \\
SAR + \textbf{COCA (ours)} & ViT-Base* & 54.5 & 55.0 & 55.9 & 56.2 & 55.3 & 61.1 & 57.7 & 58.9 & 50.9 & 71.3 & 77.2 & 68.6 & 64.5 & 70.7 & 68.0 & 61.7 \\
 & \textbullet~Combined & \textbf{56.0} & \textbf{56.6} & \textbf{57.4} & \textbf{57.5} & \textbf{56.7} & \textbf{62.9} & \textbf{59.9} & \textbf{61.9} & \textbf{53.5} & \textbf{72.3} & \textbf{78.0} & \textbf{69.5} & \textbf{66.6} & \textbf{72.0} & \textbf{69.2} & \textbf{63.3} \\ \midrule
\end{tabular}%
        }
    \end{threeparttable}
    \end{center}
    \vspace{-0.2in}
    \caption{ Experimental results on ImageNet-C (\%) show that COCA consistently outperforms the compared approaches. Furthermore, COCA can serve as a plug-and-play module, significantly enhancing the performance of existing TTA methods, such as EATA~\cite{niu2022efficient} and SAR~\cite{niu2023towards}. An asterisk (*) denotes the anchor model.}
    \label{mainres}
    \vspace{-0.15in}
\end{table*}



\paragraph{Datasets and Models} We conduct experiments mainly on the benchmark dataset, ImageNet-C~\cite{hendrycks2019benchmarking}, for test-time adaptation (TTA). This dataset contains corrupted images across 15 types of distortions in 4 main categories (noise, blur, weather, digital), with each type having 5 severity levels. Additionally, we evaluate our approach across diverse scenarios using the OfficeHome~\cite{venkateswara2017deep}, ImageNet-R~\cite{hendrycks2021many}, ImageNet-Sketch~\cite{wang2019learning}, and Cifar100-C~\cite{hendrycks2019benchmarking} datasets, where the first three represent real-world challenges. OfficeHome consists of 15,500 images of 65 classes from four domains: artistic (Ar), clipart (Cl), product (Pr), and real-world (Rw) images. ImageNet-R has renditions of 200 ImageNet classes resulting in 30,000 images, while ImageNet-Sketch consists of 50,889 images, approximately 50 images for each of the 1000 ImageNet classes. To examine the mutual promotion across different models varying in size, we select six models, (\textit{i.e.}, ViT-Large, ViT-Base, Mobile-ViT, ResNet-101, ResNet-50, and ResNet-18), which can be divided into two categories by model structure, (\textit{i.e.}, Transformer-based and Convolutional Network Networks (CNN)-based \cite{he2016deep}). The number of parameters and GMACs (Giga Multiply-Add Calculations per second) of these six models are shown in Table~\ref{params}. For reference in subsequent comparisons, we also present the results achieved using Tent~\cite{wang2020tent}, a representative baseline TTA method.

\vspace{-10pt}
\paragraph{Baseline Methods} To evaluate the effectiveness and robustness of the COCA framework, we select baseline TTA methods from three perspectives. (1) Adaptation based on entropy minimization. We include Tent~\cite{wang2020tent}, which relies solely on entropy minimization for adaptation using all test samples without incorporating collaboration. We also select DeYO~\cite{lee2024entropy2} and ROID~\cite{marsden2024universal2}. DeYO is based on entropy and a confidence metric, while ROID relies on entropy-related loss functions. (2) Integration of COCA into existing methods. We select EATA~\cite{niu2022efficient} and SAR~\cite{niu2023towards} to evaluate the performance gains achieved by embedding the COCA framework. EATA improves robustness by filtering out high-entropy samples, while SAR reduces the impact of noisy gradients with large norms that could impair adaptation. (3) Comparison with a traditional cross-model method, CoTTA~\cite{wang2022continual}, a teacher-student framework designed to mitigate error accumulation through augmentation-averaged and weight-averaged predictions.


\vspace{-13pt}
\paragraph{Implementation Details} The experiments are implemented in PyTorch and trained on an Ubuntu 20.04 system with 96 GB of memory and an Nvidia 3090 GPU. We use Stochastic Gradient Descent (SGD) with a momentum of 0.9. The batch size is 64. The learning rates for adapting ResNet-50 and ViT-Base (obtained from torchvision or timm) are set to 0.00025 and 0.001, respectively. For the trainable parameters, we follow the setup from Tent~\cite{wang2020tent}, where only the batch normalization layers are updated. For our experiments on ImageNet-C, we exclusively select severity level 5, indicating the most severe domain shifts. The implementation mechanisms of COCA as a plug-and-play module are described in Appendix~\ref{imdeatils}.

\vspace{-4pt}
\subsection{Results}
Firstly, we assess the effectiveness of COCA using two models. We report both the performance of each individually adapted model and the final outcome achieved through the co-learning process. For each model pair, an asterisk (*) denotes the anchor model, while the term ``\textbullet~Combined'' represents the overall performance resulting from cross-model co-learning. Later in this section, we will explore the applicability of COCA to more than two models. 


\begin{table}[t]
\centering
\setlength{\tabcolsep}{12pt} 
    \renewcommand{\arraystretch}{0.9} 
\resizebox{0.9\linewidth}{!}{%
\begin{tabular}{l|ccc|c}
\multicolumn{1}{c|}{Methods} & R$\to$A & R$\to$C & R$\to$P & Avg. \\ \midrule
Tent~\cite{wang2020tent} & 50.5 & 45.5 & 87.2 & 61.1 \\
EATA~\cite{niu2022efficient} & 51.4 & 46.9 & 88.4 & 62.2 \\
SAR~\cite{niu2023towards} & 50.5& 45.7& 87.6 & 61.2 \\
DeYO~\cite{lee2024entropy} & 51.8 & 47.7 & 90.5 & 63.3 \\
ROID~\cite{marsden2024universal2} & 50.9 & 48.9 & 89.3 & 63.0 \\ \midrule \midrule
COCA\textsuperscript{1} & 61.9 & 61.6 & 95.8 & 72.6 \\
COCA\textsuperscript{2} &  \textbf{65.6} & \textbf{68.3} & \textbf{97.0}& \textbf{76.9}
\end{tabular}
}
\vspace{-0.1in}
    \caption{Results on the OfficeHome dataset (\%). COCA¹ and COCA² denote collaborations between ResNet-18 and ViT-Base, and between ResNet-50 and ViT-Base, respectively.}
    \label{officehome}
    \vspace{-0.1in}
\end{table}

\vspace{-10pt}
\paragraph{Performance on ImageNet-C}
The comparative results on the dataset ImageNet-C are displayed in Table~\ref{mainres}, where ResNet-50 and ViT-Base are utilized to verify the effectiveness of our cross-model co-learning approach, COCA. From the results, we draw two key observations. 1) Consistent superiority over baselines. COCA consistently outperforms the baseline methods. Moreover, the average performance of each individual model improves, demonstrating that COCA enhances the performance of each model independently. For instance, the average performance of ViT-Base reaches 63.2\%, surpassing that of all baseline methods. 2) Significant performance gains as a plug-and-play module. COCA delivers substantial performance improvements when integrated with existing methods. For example, combining COCA with EATA~\cite{niu2022efficient} on ViT-Base achieves an impressive 69.1\%, which is significantly higher than the 62.0\% achieved by EATA alone. This improvement can be attributed to COCA’s ability to facilitate collaboration between the two models, unlocking additional optimization potential even in single-model configurations.

\vspace{-10pt}
\paragraph{Performance on More Datasets}
The results on different ImageNet models—, ImageNet-R, and ImageNet-Sketch—presented in Table~\ref{RSdatasets} (cf. Appendix), show that COCA achieves superior performance, further confirming its effectiveness in real-world domain shifts. When integrated as a plug-and-play module, COCA boosts the performance of both EATA and SAR by at least 5\%. Notably, SAR's performance ImageNet-Sketch improves by more than 10\%. More experiments on the OfficeHome and Cifar100-C datasets, as shown in Table~\ref{officehome} and Table~\ref{cifar100c} (cf. Appendix), further indicate COCA's effectiveness.

\subsection{Analysis}

\begin{table}[t]
    \begin{center}
    \begin{threeparttable}
        \resizebox{0.92\linewidth}{!}{%
            \begin{tabular}{ccc|ccc}                
                $\mathcal{L}_{sa}$ & $\mathcal{L}_{mar}$ & $\mathcal{L}_{ckd}$ & ResNet-50 & ViT-Base & \textbullet~Combined. \\
                \midrule
                \checkmark & & & 42.9 & 51.7 & - \\
                \midrule
                & \checkmark & & 47.9 & 59.9 & 62.5 \\
                
                & & \checkmark & 47.2 & 59.5 & 61.5 \\
                &\checkmark & \checkmark & 49.8 &62.1 & 64.0 \\
                \checkmark & & \checkmark & 47.2 & 61.6 & 63.8 \\
                
                \checkmark & \checkmark & & 50.3 & 61.9 & 63.8 \\
                \midrule
                \checkmark & \checkmark & \checkmark & \textbf{50.7} & \textbf{63.2} & \textbf{64.9} \\
                
            \end{tabular}%
        }
    \end{threeparttable}
    \end{center}
    \vspace{-0.24in}
    \caption{Ablation study on the loss function of COCA. Each result represents the average performance across 15 types of corruption on ImageNet-C (\%). It is evident that each loss item plays a critical role in achieving the final performance.}
    \vspace{-0.15in}
    \label{ablation}
\end{table}

\paragraph{Ablation Study on the Loss Function of COCA }
The overall loss function of COCA is composed of three components: self-adaptation ($\mathcal{L}_{sa}$), cross-model knowledge distillation ($\mathcal{L}_{ckd}$), and marginal entropy minimization ($\mathcal{L}_{mar}$). An ablation study on these components is presented in Table~\ref{ablation}, highlighting the importance of each to COCA's performance. When evaluating the individual contributions of these components, both $\mathcal{L}_{mar}$ and $\mathcal{L}_{ckd}$ achieve higher performance compared to $\mathcal{L}_{sa}$, suggesting that collaborative knowledge provides more effective learning signals to the models. Notably, when only the self-adaptation entropy loss ($\mathcal{L}_{sa}$) is applied, COCA is equivalent to Tent~\cite{wang2020tent}. For this reason, the combined result is denoted as ``-” in the table.

\begin{figure}[t]
\centering
    \includegraphics[width=0.85\linewidth]{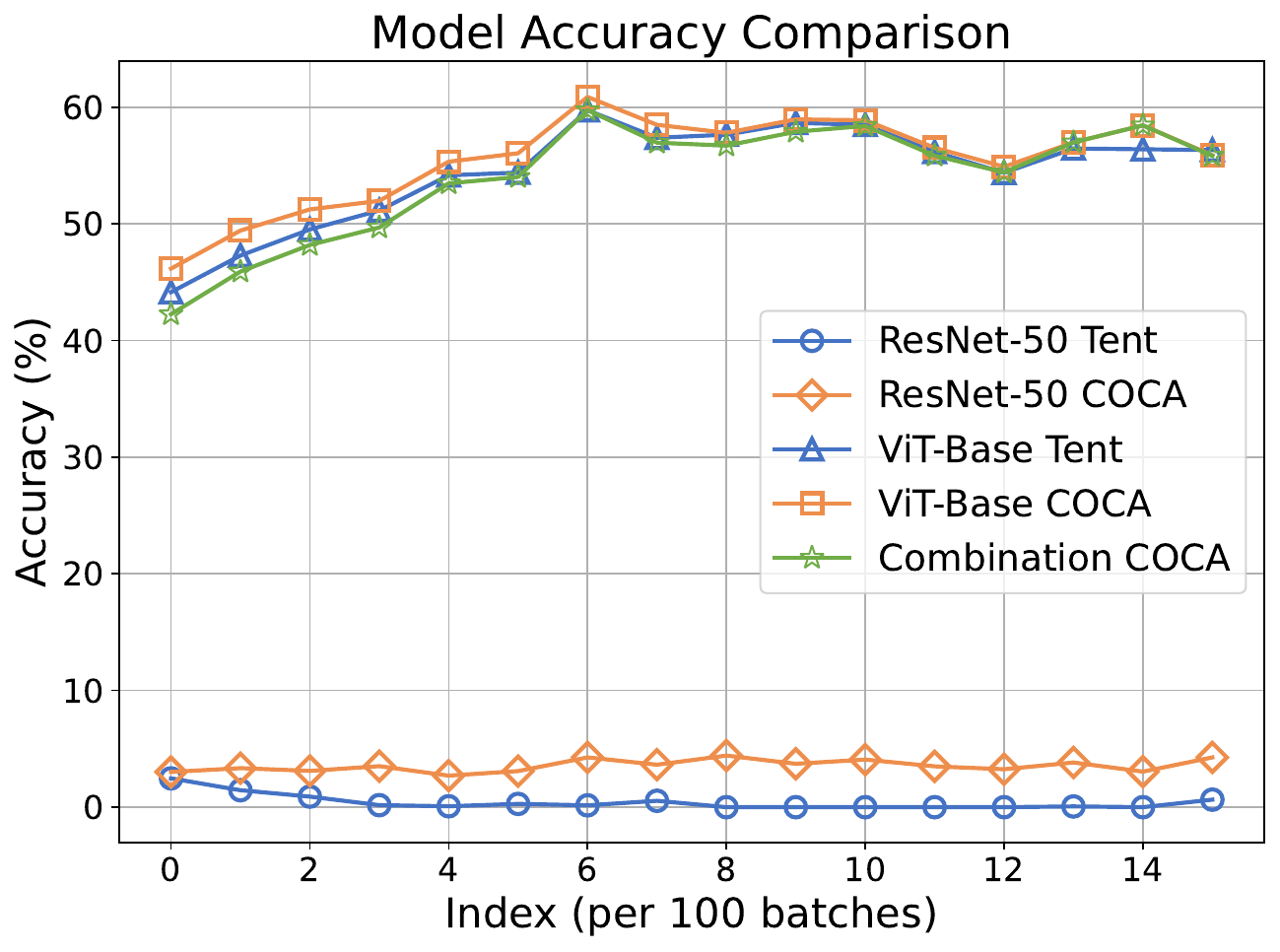}
    \vspace{-0.12in}
    \caption{Robustness of COCA stems from $\tau$. In the label-shift scenario~\cite{niu2023towards}, COCA maintains high performance even when one of the models—such as ResNet-50 in this figure—collapses. The evaluation is conducted on ImageNet-C with Gaussian noise (\%).} 
\label{lsrobust}
\vspace{-0.07in}
\end{figure}

\begin{table}[t]
    \centering
    \setlength{\tabcolsep}{9pt} 
    \renewcommand{\arraystretch}{1} 
    \begin{threeparttable} 
    \resizebox{0.95\linewidth}{!}{%
        \begin{tabular}{c|c|cc}
            Models & \multicolumn{1}{l|}{Accuracy (\%)} & Parameters (M) & GMACs \\
            \midrule
            ResNet-18 & 35.1 & 11.7 & 1.8 \\
            ResNet-50 & 42.9 & 25.6 & 4.1 \\
            ResNet-101 & 46.5 & 44.5 & 7.8 \\
            \midrule
            Mobile-ViT & 41.0 & 10.6 & 4.1 \\
            ViT-Base & 51.7 & 86.6 & 16.9 \\
            ViT-Large & 67.8 & 304.2 & 77.8 \\
        \end{tabular}%
    }
    \vspace{-0.1in}
    \caption{Comparison of Model Profiles: performance (\%), the number of parameters (in millions), and GMACs. The performance is the average adaptation performance across 15 types of corruption on ImageNet-C, evaluated using Tent ~\cite{wang2020tent}.}
    \label{params}
    \vspace{-0.15in} 
    \end{threeparttable}
\end{table}

\vspace{-5pt}
\paragraph{The Learnable Parameter $\tau$ Enhances the Robustness of COCA}


As shown in Fig.~\ref{Tcompare} of section.\ref{sec:Method Section}, it is necessary to introduce a learnable parameter $\tau$, which can enhance the robustness of COCA. In certain scenarios, such as label shifts~\cite{niu2023towards}, the ResNet-50-BN model may be prone to collapse. However, in these cases, the collaborative model, ViT-Base, continues to learn effectively, as illustrated in Fig.~\ref{lsrobust}. The accuracy of the ResNet-50-BN model approaches zero, indicating that its predictions diverge significantly from those of ViT. Under such conditions, the value of $\tau$ increases substantially to balance the outputs of the two models. Consequently, the contribution of ResNet-50 to the collaborative knowledge computation diminishes, as shown in Eq.~\ref{marginalent}, thereby preventing any harmful effects on collaboration. This ensures that the final performance remains close to that of ViT-Base under COCA. Therefore, if one model collapses at test time, we can discard it and rely solely on the adapted predictions of the other model to maintain performance. In our experiments, the learnable parameter $\tau$ ranges from 1 to 5. The Appendix~\ref{VisualizeCOCA} further visualizes the benefits of COCA, highlighting how $\tau$ enables robust co-adaptation.

\vspace{-10pt}

\paragraph{Influence of Anchor Model Selection}
The influence of selecting the anchor model is shown in Table~\ref{anchorexchange}, where ``Anc." and ``Aux." refer to the accuracy of the anchor model and the auxiliary model, respectively. We observe that choosing any model as the anchor improves the performance compared to single-model TTA. Additionally, selecting a larger model as the anchor leads to better performance. Notably, the performance gap between the two models increases as the difference in the number of parameters grows. For example, when selecting ViT-Base as the anchor model, the performance is higher than when the auxiliary model is Mobile-ViT, which has fewer parameters than ResNet-50.
\begin{table}[t]
    \begin{center}
    \setlength{\tabcolsep}{9pt} 
    \renewcommand{\arraystretch}{1.1} 
    \begin{threeparttable}
        \resizebox{\linewidth}{!}{%
            \begin{tabular}{cc|ccc}
Anchor & Auxiliary & Anc. & Aux. & \textbullet~Combined. \\ 
\midrule
ResNet-18 & ResNet-50 & 37.3 & 43.4 & 43.8 \\
ResNet-50 & ResNet-18 & 44.0 & 38.9 & \textbf{45.6} \\ \midrule
ResNet-50 & ViT-Base & 48.5 & 63.0 & 62.2 \\
ViT-Base & ResNet-50 & 63.2&50.7& \textbf{64.9} \\ \midrule
Mobile-ViT & ViT-Base & 47.7 & 63.3 & 61.6 \\
ViT-Base & Mobile-ViT & 64.0& 47.5& \textbf{64.5}
\end{tabular}%
        }
    \end{threeparttable}
    \end{center}
    \vspace{-0.25in}
    \caption{Influence of anchor model selection. Each result represents the average performance across 15 types of corruption on ImageNet-C (\%). It is clear that selecting the larger model as the anchor leads to higher performance.}
    \label{anchorexchange}
    \vspace{-0.15in}
\end{table}

\vspace{-10pt}

\paragraph{Performance Across Different Model Pairs \& Influence of Models' Parameters and Architectures} We first examine the performance of six models under a single-model adaptation framework following Tent~\cite{wang2020tent}, with the results shown in Table~\ref{params}. Next, we form various pairs from these models to further evaluate the applicability of COCA. As presented in Table~\ref{allmodelssup} (cf. Appendix), COCA consistently achieves robust performance across all model pairs, demonstrating that the cross-model co-learning mechanism remains effective even when the paired models share the same architecture, as in the case of ViT-Base and Mobile-ViT. Additionally, under COCA, each model consistently outperforms single-model adaptation following Tent~\cite{wang2020tent}.

From these comparative results, we derive two key insights: (1) models with a larger number of parameters tend to exhibit enhanced performance, and (2) architectural diversity offers additional benefits. Further analysis is provided in Appendix~\ref{MoreAcross}.

\vspace{-10pt}

\paragraph{Computational Complexity Analysis} The comparative results in Table~\ref{eff} show that COCA requires similar computational time to Tent~\cite{wang2020tent} and EATA~\cite{niu2022efficient}, while being more efficient than SAR~\cite{niu2023towards} and CoTTA~\cite{wang2022continual}. Notably, unlike CoTTA, which relies on a teacher-student framework, COCA demands less memory and GPU time. This makes COCA particularly useful when GPU resources are limited. In such cases, a smaller model like ResNet-18 can be used in place of a larger model such as ResNet-50. Although the performance may slightly decrease compared to the larger model, it can still yield satisfactory results.

\begin{table}[t]
    \begin{center}
    \setlength{\tabcolsep}{3pt} 
    \renewcommand{\arraystretch}{1} 
    \begin{threeparttable}
        \resizebox{\linewidth}{!}{%
            \begin{tabular}{l|c|cc}
\multicolumn{1}{c|}{Methods} & Accuracy (\%) & GPU Time (s) & Memory Usage \\ \midrule
Source Only & 35.0 & 115 & 1300 \\ \midrule
CoTTA~\cite{wang2022continual} & 56.7 & 724 & 17,211 \\
Tent~\cite{wang2020tent} & 51.7 & 234 & 5202 \\
SAR~\cite{niu2023towards} & 51.9 & 472 & 5532 \\
EATA~\cite{niu2022efficient} & 55.6 & 246 & 5202 \\
DeYO~\cite{lee2024entropy} & 54.1 & 318 & 5371 \\ 
ROID~\cite{marsden2024universal2} & 52.0 & 371 & 7972 \\ \midrule
\textbf{COCA\textsuperscript{1} (ours)}& 56.8 & 315 & 6574 \\
\textbf{COCA\textsuperscript{1} (ours)} +~EATA & \textbf{59.7} & 316 & 6574 \\ \midrule
\textbf{COCA\textsuperscript{2} (ours)}& 58.3 & 317 & 10,531 \\
\textbf{COCA\textsuperscript{2} (ours)} +~EATA & \textbf{60.9} & 319 & 10,531

\end{tabular}%
        }
    \end{threeparttable}
    \end{center}
    \vspace{-0.23in}
    \caption{Computational complexity analysis. COCA¹ and COCA² denote the collaborations between ResNet-18 and ViT-Base, and between ResNet-50 and ViT-Base, respectively. The results are based on ImageNet-C (Gaussian noise).}
    \label{eff}
    \vspace{-0.15in}
\end{table}

\vspace{-7pt}

\paragraph{Rethinking the Advantages of COCA} The consistent superiority of COCA can be attributed to the fact that, when dealing with out-of-distribution data, the two models leverage their unique strengths to collaboratively achieve robust performance. COCA serves as a general framework for cross-model co-learning, allowing a smaller model like MobileViT to effectively guide a larger model like ViT-Base. Furthermore, the co-adaptation between edge models like ResNet-18 and MobileViT surpasses the performance of much larger models such as ResNet-101 (Table~\ref{params}, Table~\ref{allmodelssup} in Appendix), offering valuable insights for tackling resource-constrained tasks. Additionally, COCA updates the parameters of both models simultaneously without causing a significant increase in GPU time (Table~\ref{eff}). 
\vspace{-7pt}
\paragraph{Is COCA Applicable to More Than Two Models?}
Building on COCA’s mechanism, a new question emerges: Is COCA applicable to multiple models? In particular, can incorporating a third small model further improve the TTA performance? To answer this, we verify our effectiveness across multiple models in Appendix~\ref{suppl:sec:three-models}, which showcases COCA's consistent improvement in overall accuracy with more collaborating models, underscoring our scalability. Results and analyses are detailed in Appendix~\ref{suppl:sec:three-models}.

\vspace{6pt}
\section{Conclusions}
In this paper, we aim to investigate the co-learning mechanism to enhance test-time adaptation (TTA). Notably, we uncover two key insights regarding different models: 1) They provide complementary knowledge from training, and 2) They exhibit varying robustness against noisy learning signals during testing. To this end, we propose COCA, a co-learning TTA approach that promotes bidirectional cooperation across models. COCA integrates complementary knowledge to mitigate biases and enables tailored adaptation strategies for each model’s strengths. Notably, a learnable temperature, $\tau$, is introduced to align the outputs from models of different sizes. Extensive experiments validate COCA's effectiveness across diverse models and offer new insights into TTA. For example, we show that a smaller model like ResNet-18, attributed to its robustness to noise at testing, can significantly enhance the TTA of a much larger model like ViT-large. 
Additionally, COCA can also serve as a plug-and-play module that seamlessly complements traditional TTA methods for more effective and robust solutions. 

{
    \small
    \bibliographystyle{ieeenat_fullname}
    \bibliography{main}
}

\clearpage
\setcounter{page}{1}
\maketitlesupplementary
\appendix
\section{Pseudo-code of COCA}
\label{pscode}
\begin{algorithm}
\begin{spacing}{1.1}
\caption{The pipeline of proposed COCA.}
\label{alg:coca}
\textbf{Input:} Test samples $D^{\text{test}}=\{x_j\}_{j=1}^M$, the pre-trained models $f_{\theta_{1}}, f_{\theta_{2}}$ with trainable parameters $\widetilde{\theta}_{1}, \widetilde{\theta}_{2} \subseteq  \theta_{1}, \theta_{2}$, steps $K$ for learnable $\tau$.
\begin{algorithmic}[1]

    \STATE Initialize $\tau = 1 $
    \FOR{each batch of test samples }
        \STATE Calculate predictions $\hat{y}_{\theta_{1}}$ $\hat{y}_{\theta_{2}}$ via ${f}
        _{\theta_{1}}$ and ${f}_{\theta_{2}}$
        \FOR{ $i=1 $ \textbf{to} $ K$}
            \STATE Update $\tau$ via Eqn.(\ref{tloss})
        \ENDFOR
        \STATE Calculate combined predictions $\hat{y}_{e}$ via Eqn.(\ref{tau_ensemble})
        \STATE Calculate co-adaptation loss $\mathcal{L}_{col}$ via Eqn.(\ref{colloss})
        \STATE Calculate self-adaptation loss $\mathcal{L}_{sa}$ via Eqn.(\ref{selfloss})
        \STATE Update $\widetilde{\theta}_{1}$ and $\widetilde{\theta}_{2}$ via Eqn.(\ref{fullloss})
        
    \ENDFOR 
\end{algorithmic}
\textbf{Output:} Predictions $\{ \hat{y}^j_{\theta_{1}}, \hat{y}^j_{\theta_{2}},\hat{y}^j_e\}_{j=1}^M $ for all $x_{j} \in D^{\text{test}}$.
\end{spacing}
\end{algorithm}


\section{Implementation Details of COCA Integrated with Baseline Methods}
\label{imdeatils}
As discussed in the main manuscript, COCA can also serve as a plug-and-play module to enhance existing TTA methods by applying its loss function (Eq.~\ref{fullloss}). For instance, \textit{EATA~\cite{niu2022efficient} + COCA} utilizes EATA's sample filtering and weighting mechanism to compute both the self-adaptation loss for individual models and the co-adaptation loss for selected model pairs. Similarly, \textit{SAR~\cite{niu2023towards} + COCA} employs the Sharpness-Aware Minimization (SAM) optimizer from SAR's original formulation to jointly optimize model parameters while integrating its entropy-based reliability filtering mechanism. Moreover, the implementation details of all the baseline methods follow their original implementation formulations.

\section{The Study of Step $K$ for Updating the Learnable Parameter $\tau$}
\label{kstudy}
In Section~\ref{sec:Method Section} of the main paper, we introduced the temperature-scaled parameter $\tau$  to facilitate the combination of two models. We now examine how to determine the optimal value of $\tau$, which is governed by the parameter $K$ (the number of updating steps). The results, shown in Table~\ref{tepoch}, indicate that only a small number of epochs—approximately 3 to 5—are needed to achieve the optimal value of $\tau$. Furthermore, these results demonstrate that COCA is not highly sensitive to the choice of $K$. Note that we apply stop gradients to the model outputs and update only the parameter $\tau$, without performing backpropagation. As a result, learning $\tau$ incurs negligible computational cost, even with a large number of update steps $K$, ensuring efficient adaptation.

\section{Visualizing the Advantages of COCA}
\label{VisualizeCOCA}
\begin{figure*}[t]
\centering
    \includegraphics[width=0.93\linewidth]{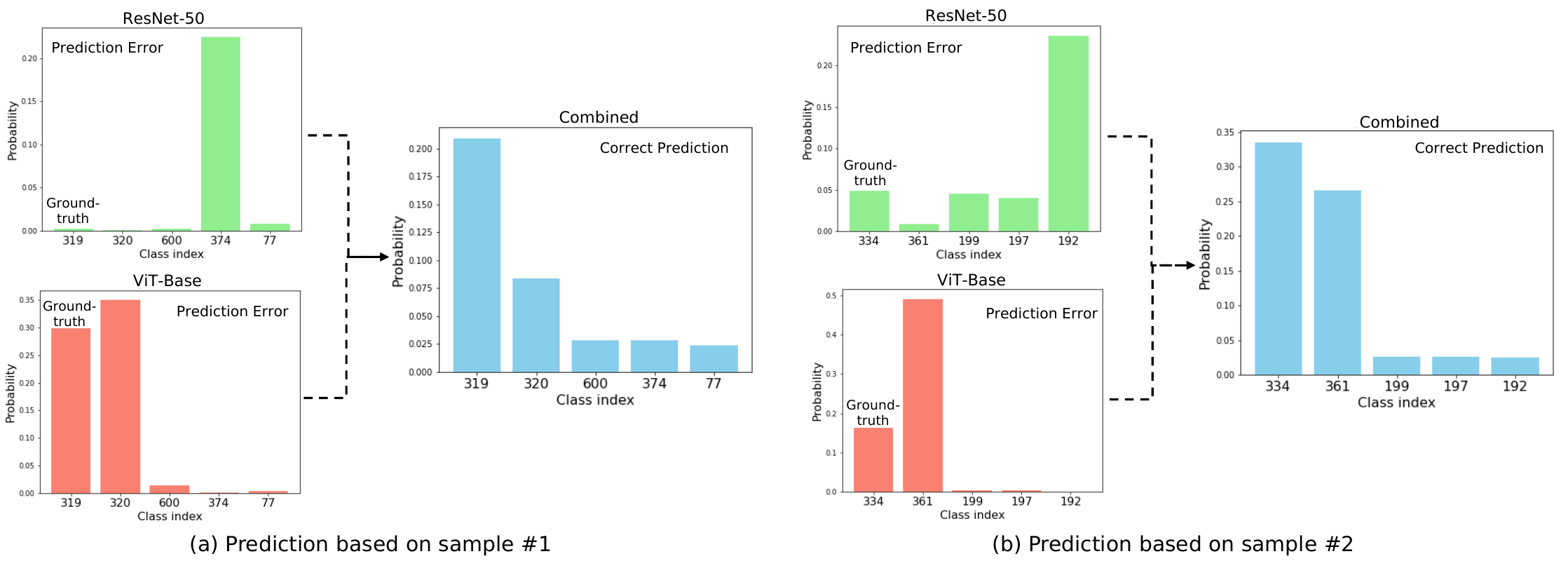}
    \vspace{-0.1in}
    \caption{A sample-level analysis highlights the advantages of our proposed cross-model co-learning approach. For each sample, we select the top five predicted probabilities from both ResNet-50 and ViT-Base. However, the final correct prediction, determined by COCA, differs from each of the initial predictions.}
    \vspace{-0.1in}
\label{samplelevel}
\end{figure*}

Extensive comparative results demonstrate that COCA consistently outperforms a single model. To further illustrate this advantage, we present two sample cases where, even if individual models make incorrect predictions, their combined output can still yield the correct result. Specifically, we examine Sample \#1 and Sample \#2 to showcase COCA's benefits. As shown in Fig.\ref{samplelevel}, both the ViT-Base and ResNet-50 models confidently produce incorrect predictions, with the assigned probability for the predicted category significantly higher than for the others. In contrast, our proposed approach, COCA, generates entirely different but correct predictions. Additionally, we use these two samples to explain why COCA can make accurate predictions, leveraging the learnable parameter $\tau$, as detailed in Table~\ref{Sampleleveltable}. The combined logit is the average of the logits from ViT-base and the logits from ResNet-50, where the latter is scaled by a factor of $\tau$. Take the predicted logits for category 192 as an example: $5.7 = (1.6 + 9.8) / 2$.

\begin{table}[t]
     \setlength{\tabcolsep}{6pt} 
    \renewcommand{\arraystretch}{1.1} 
    \begin{center}
    \begin{threeparttable}
        \resizebox{0.95\linewidth}{!}{%
            \begin{tabular}{l|cc|c}
                \multicolumn{1}{c|}{Methods} & ImageNet-R & ImageNet-Sketch & Avg.  \\
                \midrule
                Tent~\cite{wang2020tent} & 54.2 & 32.5 & 43.4 \\
                CoTTA~\cite{wang2022continual} & 56.6 & 42.4 & 49.5 \\
                \textbf{COCA (ours)} & \textbf{57.7} & \textbf{44.0} & \textbf{50.9} \\
                \midrule
                EATA~\cite{niu2022efficient} & 55.4 & 40.0 & 47.7 \\
                EATA +~\textbf{COCA (ours)} & \textbf{63.8} & \textbf{47.8} & \textbf{55.8} \\
                \midrule
                SAR~\cite{niu2023towards} & 55.0 & 33.8 & 44.4 \\
                SAR +~\textbf{COCA (ours)} & \textbf{60.7} & \textbf{44.4} & \textbf{52.6} \\
            \end{tabular}%
        }
    \end{threeparttable}
    \end{center}
    \vspace{-0.25in}
    \caption{Comparative results (\%) on the real-world datasets ImageNet-R and ImageNet-Sketch show that COCA consistently outperforms all baseline methods. Additionally, COCA functions as a plug-and-play module that significantly enhances entropy-based TTA methods, such as EATA~\cite{niu2022efficient} and SAR~\cite{niu2023towards}.}
    \vspace{-0.15in}
    \label{RSdatasets}

\end{table}

\section{Ratio Between Co-adaptation and Self-adaptation}
\label{Ratio}
COCA is designed to foster bidirectional improvement throughout the entire adaptation process. It can also be seamlessly integrated as a plug-and-play module to boost existing Test-Time Adaptation (TTA) methods. To keep our approach as simple as possible, we minimize the number of hyper-parameters, initially setting the co-adaptation to self-adaptation ratio to 1:1 as indicated in Eq.~\ref{fullloss}. Further experiments with \textit{EATA+COCA (ours)} on ImageNet-C reveal that a ratio of 1:2 slightly enhances accuracy—from 69.1\% to 69.5\%. Therefore, we view the 1:1 setting as a balanced compromise, offering faster computation with a modest performance trade-off compared to the marginal gains achieved with a higher computational cost.

\begin{table}[t]
\setlength{\tabcolsep}{11pt} 
    \renewcommand{\arraystretch}{0.9} 
\resizebox{0.95\linewidth}{!}{%
\begin{tabular}{c|ccc|c}

\multicolumn{5}{c}{Sample \#1}\\ \midrule \midrule
Model \textbackslash Class & 98 & 146 & 356 & Predicted class \\ \midrule
ResNet-50 & 5.4 & 2.1 & 7.5 & 356 (\ding{55}) \\
ViT-Base & 5.9 & 8.6 & 1.6 & 146 (\ding{55}) \\
ResNet-50 / $\tau$ & 8.1 & 5.2 & 9.0 & - \\ \midrule
Combined & 7.0 & 6.9 & 5.3 & 98 (\checkmark)\\ \midrule 
\multicolumn{5}{c}{Sample \#2} \\ \midrule \midrule
Model\textbackslash Class & 192 & 334 & 361 & Predicted class \\ \midrule
ResNet-50 & 8.0 & 6.5 & 4.8 & 192 (\ding{55}) \\
ViT-Base & 1.6 & 6.9 & 8.0 & 361 (\ding{55}) \\
ResNet-50 / $\tau$ & 9.8 & 9.7 & 8.2 & - \\ \midrule
Combined & 5.7 & 8.3 & 8.1 & 334 (\checkmark)
\end{tabular}%
}
\vspace{-0.1in}
\caption{Further explanation of Fig.~\ref{samplelevel} on how COCA makes accurate predictions for these two samples. The values, all less than 10, represent the predicted logits. }
\label{Sampleleveltable}
\vspace{-0.2in}
\end{table}

\section{The Influence of Model Weights}
\label{weights}
COCA remains effective even when the two models share the same deep architecture and differ only in their pre-trained weights. Table~\ref{samemodel} presents results based on ResNet-18 and ResNet-50, demonstrating that the cross-model co-learning mechanism remains robust. In this table, the superscript \textsuperscript{1} indicates that the model was pre-trained on the Instagram-1B hashtag dataset using semi-weakly supervised learning and fine-tuned on ImageNet~\cite{he2016deep}. Models without this superscript were initialized using the pre-trained weights provided by PyTorch. Notably, unlike previous findings, the performance of the anchor model does not consistently outperform that of the auxiliary model.

\begin{table}[t]
    \begin{center}
    \setlength{\tabcolsep}{9pt} 
    \renewcommand{\arraystretch}{1.1} 
    \begin{threeparttable}
        \resizebox{\linewidth}{!}{
            \begin{tabular}{cc|ccc}
Anchor & Auxiliary & Anc. & Aux & \textbullet~Combined \\ \midrule
ResNet-18\textsuperscript{1} & ResNet-18 & 31.2 & 37.4 & 38.6 \\
ResNet-18 & ResNet-18\textsuperscript{1} & 37.2 & 31.0 & \textbf{39.9} \\ \midrule
ResNet-50\textsuperscript{1} & ResNet-50 & 41.7 & 45.8 & 48.5 \\
ResNet-50 & ResNet-50\textsuperscript{1} & 45.7 & 41.1 & \textbf{49.2}
\end{tabular}%
        }
    \end{threeparttable}
    \end{center}
    \vspace{-0.23in}
    \caption{Investigating whether COCA continues to deliver performance improvements when two models share the same deep architecture but differ in their pre-trained weights. Each result represents the average performance across 15 types of corruption on ImageNet-C (\%). For each pair, the two models have different pre-trained weights. }
    \label{samemodel}
    \vspace{-0.15in}
\end{table}


\begin{table*}[t]
    \setlength{\tabcolsep}{3pt} 
    \renewcommand{\arraystretch}{1.1} 
\resizebox{1.0\linewidth}{!}{%
\begin{tabular}{lccccccccccccccccc}
\multicolumn{1}{c}{} &  & \multicolumn{3}{c}{Noise} & \multicolumn{4}{c}{Blur} & \multicolumn{4}{c}{Weather} & \multicolumn{4}{c}{Digital} &  \\ \midrule
\multicolumn{1}{c|}{Methods} & \multicolumn{1}{c|}{Models} & Gauss & Shot & \multicolumn{1}{c|}{Impul} & Defoc & Glass & Motion & \multicolumn{1}{c|}{Zoom} & Snow & Frost & Fog & \multicolumn{1}{c|}{Brit} & Contr & Elastic & Pixel & \multicolumn{1}{c|}{JPEG} & Avg. \\ \midrule
\multicolumn{1}{l|}{\multirow{2}{*}{Source Only}} & \multicolumn{1}{c|}{ResNet-50} & 4.1 & 4.8 & \multicolumn{1}{c|}{2.0} & 26.1 & 11.2 & 26.1 & \multicolumn{1}{c|}{31.3} & 6.9 & 14.4 & 40.4 & \multicolumn{1}{c|}{9.6} & 36.9 & 20.1 & 11.5 & \multicolumn{1}{c|}{13.6} & 17.3 \\
\multicolumn{1}{l|}{} & \multicolumn{1}{c|}{ViT-Base} & 24.5 & 28.7 & \multicolumn{1}{c|}{29.4} & 59.6 & 23.3 & 53.1 & \multicolumn{1}{c|}{63.6} & 56.9 & 57.7 & 67.2 & \multicolumn{1}{c|}{70.6} & 76.2 & 45.3 & 36.2 & \multicolumn{1}{c|}{50.5} & 49.5 \\ \midrule
\multicolumn{1}{l|}{\multirow{2}{*}{Tent~\cite{wang2020tent}}} & \multicolumn{1}{c|}{ResNet-50} & 32.4 & 34.9 & \multicolumn{1}{c|}{33.2} & 64.9 & 38.7 & 61.0 & \multicolumn{1}{c|}{67.7} & 59.5 & 58.8 & 61.0 & \multicolumn{1}{c|}{73.3} & 67.0 & 51.8 & 59.7 & \multicolumn{1}{c|}{43.6} & 53.8 \\
\multicolumn{1}{l|}{} & \multicolumn{1}{c|}{ViT-Base} & 49.9 & 52.6 & \multicolumn{1}{c|}{56.1} & 76.5 & 32.5 & 71.5 & \multicolumn{1}{c|}{77.9} & 77.7 & 76.1 & 71.8 & \multicolumn{1}{c|}{88.9} & 74.3 & 58.5 & 42.8 & \multicolumn{1}{c|}{66.3} & 64.8 \\ \midrule
\multicolumn{1}{l|}{\multirow{2}{*}{EATA~\cite{niu2022efficient}}} & \multicolumn{1}{c|}{ResNet-50} & 35.5 & 37.4 & \multicolumn{1}{c|}{36.9} & 33.5 & 32.9 & 46.8 & \multicolumn{1}{c|}{52.5} & 51.6 & 45.8 & 60.0 & \multicolumn{1}{c|}{68.6} & 42.4 & 58.0 & 60.9 & \multicolumn{1}{c|}{55.5} & 55.4 \\
\multicolumn{1}{l|}{} & \multicolumn{1}{c|}{ViT-Base} & 54.1 & 54.8 & \multicolumn{1}{c|}{55.0} & 54.0 & 54.6 & 61.6 & \multicolumn{1}{c|}{57.8} & 63.5 & 62.8 & 71.3 & \multicolumn{1}{c|}{77.0} & 66.8 & 64.6 & 71.4 & \multicolumn{1}{c|}{68.1} & 66.7 \\ \midrule
\multicolumn{1}{l|}{\multirow{2}{*}{ROID~\cite{marsden2024universal2}}} & \multicolumn{1}{c|}{ResNet-50} & 36.2 & 39.5 & \multicolumn{1}{c|}{33.4} & 66.2 & 39.3 & 63.6 & \multicolumn{1}{c|}{68.0} & 62.5 & 62.2 & 64.8 & \multicolumn{1}{c|}{75.8} & 70.4 & 51.7 & 58.4 & \multicolumn{1}{c|}{42.4} & 55.6 \\
\multicolumn{1}{l|}{} & \multicolumn{1}{c|}{ViT-Base} & 55.6 & 57.9 & \multicolumn{1}{c|}{61.2} & 75.2 & 48.1 & 71.9 & \multicolumn{1}{c|}{77.1} & 73.3 & 75.5 & 76.4 & \multicolumn{1}{c|}{83.7} & 84.2 & 64.5 & 68.6 & \multicolumn{1}{c|}{64.4} & 69.2 \\ \midrule
\multicolumn{1}{l|}{\multirow{2}{*}{DeYO~\cite{lee2024entropy}}} & \multicolumn{1}{c|}{ResNet-50} & 35.9 & 39.4 & \multicolumn{1}{c|}{32.3} & 66.7 & 39.3 & 63.4 & \multicolumn{1}{c|}{68.1} & 62.4 & 61.7 & 65.1 & \multicolumn{1}{c|}{75.7} & 71.3 & 51.8 & 58.2 & \multicolumn{1}{c|}{42.2} & 55.6 \\
\multicolumn{1}{l|}{} & \multicolumn{1}{c|}{ViT-Base} & 49.6 & 52.7 & \multicolumn{1}{c|}{58.4} & 76.0 & 42.8 & 72.8 & \multicolumn{1}{c|}{76.7} & 73.9 & 74.3 & 75.9 & \multicolumn{1}{c|}{85.0} & 83.1 & 62.0 & 62.5 & \multicolumn{1}{c|}{62.6} & 67.2 \\ \midrule
\multicolumn{1}{l|}{\textbf{COCA\textsuperscript{1} (ours)}} & \multicolumn{1}{c|}{Combined} & 52.3 & 48.4 & \multicolumn{1}{c|}{65.6} & 81.2 & 55.1 & 79.5 & \multicolumn{1}{c|}{82.4} & 79.7 & 79.2 & 79.9 & \multicolumn{1}{c|}{88.2} & 87.2 & 69.1 & 77.1 & \multicolumn{1}{c|}{68.6} & 72.9 \\
\multicolumn{1}{l|}{\textbf{COCA\textsuperscript{2} (ours)}} & \multicolumn{1}{c|}{Combined} & \textbf{62.1} & \textbf{64.5} & \multicolumn{1}{c|}{\textbf{70.0}} & \textbf{82.4} & \textbf{61.8} & \textbf{80.1} & \multicolumn{1}{c|}{\textbf{82.3}} & \textbf{79.9} & \textbf{80.8} & \textbf{81.0} & \multicolumn{1}{c|}{\textbf{88.5}} & \textbf{87.5} & \textbf{71.9} & \textbf{77.6} & \multicolumn{1}{c|}{\textbf{69.4}} & \textbf{76.0}\\ \midrule
\end{tabular}
}
\vspace{-0.1in}
    \caption{Comparative results based on the Cifar100-C dataset (\%). COCA¹ and COCA² refer to the collaborations between ResNet-18 and ViT-Base, and between ResNet-50 and ViT-Base, respectively.}
    \label{cifar100c}

\end{table*}

\begin{table*}[t]
    \setlength{\tabcolsep}{8pt} 
    \renewcommand{\arraystretch}{1} 
    
    \begin{center}
    \begin{threeparttable}
    \resizebox{1\linewidth}{!}{%
\begin{tabular}{c|ccccccccccc}
Model / Steps & $K=0$ & $K=1$ & $K=2$ & $K=3$ & $K=4$ & $K=5$ & $K=6$ & $K=7$ & $K=8$ & $K=9$ & $K=10$ \\ \midrule
ResNet-18 & 27.6 & 29.5 & 29.5 & 29.4 & 29.4 & 29.5 & 29.2 & 29.5 & 29.4 & 29.0 & 29.2 \\
ResNet-50* & 32.1 & 32.8 & 32.6 & 32.6 & 32.6 & 32.8 & 32.4 & 32.6 & 32.6 & 32.5 & 32.5 \\
\textbullet~Combined & 33.3 & 35.2 & 35.1 & 35.1 & 35.1 & 35.2 & 34.8 & 35.1 & 35.1 & 35.0 & 35.0 \\ \midrule
Mobile-ViT & 32.1 & 32.5 & 32.7 & 32.2 & 32.5 & 32.1 & 32.5 & 32.3 & 32.4 & 32.4 & 32.5 \\
ViT-Base* & 54.8 & 55.5 & 55.5 & 55.6 & 55.6 & 55.6 & 55.6 & 55.6 & 55.5 & 55.5 & 55.6 \\
\textbullet~Combined & 52.9 & 54.9 & 55.1 & 55.2 & 55.4 & 55.4 & 55.4 & 55.5 & 55.4 & 55.3 & 55.4 \\ \midrule
Mobile-ViT & 29.1 & 30.1 & 30.3 & 30.3 & 30.2 & 30.3 & 29.2 & 30.3 & 30.3 & 30.4 & 30.0 \\
ResNet-50* & 34.2 & 34.3 & 34.3 & 34.4 & 34.3 & 34.4 & 34.3 & 34.4 & 34.4 & 34.4 & 34.4 \\
\textbullet~Combined & 36.9 & 37.7 & 37.9 & 37.9 & 37.9 & 38.0 & 37.3 & 38.0 & 38.0 & 38.0 & 37.9 \\ \midrule
ResNet-50 & 40.9 & 41.8 & 41,5 & 41,5 & 41,5 & 41.5 & 41.6 & 41.6 & 41.6 & 41.6 & 41.6 \\
ViT-Base & 56.0 & 56.3 & 56.2 & 56.2 & 56.2 & 56.2 & 56.2 & 56.2 & 56.2 & 56.2 & 56.2 \\
\textbullet~Combined & 57.3 & 58.1 & 58.0 & 58.0 & 58.0 & 58.0 & 58.0 & 58.0 & 58.0 & 58.0 & 58.0\\ \midrule
\end{tabular}%
 }

    \end{threeparttable}
    \end{center}
    \vspace{-0.2in}
    \caption{Impact of the parameter update iterations, $K$, for $\tau$ on cross-model collaborative learning performance (\%). An asterisk (*) denotes the anchor model. This experiment is also based on ImageNet-C (Gaussian Noise under severity level 5). $K=0$ represents that the outputs of two pre-trained models are combined by averaging, without learning $\tau$. }
    \label{tepoch}
    \vspace{-0.1in}
\end{table*}


\section{More Analysis on the Influence of Model Parameters and Architectures}
\label{MoreAcross}

Two key insights emerge from the extensive comparative results: 1) More parameters enhance performance. Increasing the parameter count in the auxiliary model enhances overall performance when the anchor model is fixed. For instance, the accuracy improves from 64.9\% with ResNet-50 and ViT-Base to 67.1\% with ResNet-101 and ViT-Base. This improvement is likely because models with more parameters and the same deep architecture can facilitate more comprehensive learning. 2) Architectural diversity offers benefits. Diversity in deep architectures between models improves performance. For pairs of models sharing one common model, the pair with differing architectures outperforms the one with similar parameter sizes but identical architectures. For example, the accuracy of ResNet-18 and ResNet-50 is 45.6\%, while Mobile-ViT and ResNet-50 achieve 50.5\%. This advantage may arise because distinct architectures enable the models to learn more diverse representations from the same source domain.

\begin{table*}[t]
    \setlength{\tabcolsep}{3pt} 
    \renewcommand{\arraystretch}{1.03} 
    \begin{center}
    \begin{threeparttable}
        \resizebox{0.95\linewidth}{!}{%
    \begin{tabular}{c|ccc|cccc|cccc|cccc|c}
 \multicolumn{1}{c}{}& \multicolumn{3}{c}{Noise} & \multicolumn{4}{c}{Blur} & \multicolumn{4}{c}{Weather} & \multicolumn{4}{c}{Digital} &  \\ \midrule
Models & Gauss & Shot & Impul & Defoc & Glass & Motion & Zoom & Snow & Frost & Fog & Brit & Contr & Elastic & Pixel & JPEG & Avg. \\ \midrule
Mobile-ViT & 29.6 & 32.5 & 32.9 & 30.2 & 29.3 & 44.5 & 50.7 & 52.0 & 47.7 & 60.4 & 70.1 & 41.9 & 54.1 & 54.7 & 52.3 & { 45.5} \\
ResNet-18* & 28.5 & 30.3 & 29.0 & 26.6 & 26.8 & 37.6 & 43.8 & 41.9 & 37.3 & 51.2 & 60.2 & 34.2 & 49.3 & 52.5 & 48.1 & { 39.8} \\
\textbullet~Combined & \textbf{34.9} & \textbf{37.3} & \textbf{36.7} & \textbf{33.8} & \textbf{33.0} & \textbf{47.6} & \textbf{53.3} & \textbf{53.7} & \textbf{48.7} & \textbf{61.7} & \textbf{70.4} & \textbf{43.9} & \textbf{57.6} & \textbf{59.2} & \textbf{56.0} & { \textbf{48.5}} \\ \midrule
Mobile-ViT & 29.7 & 29.0 & 33.2 & 30.1 & 26.5 & 44.8 & 50.9 & 51.3 & 48.0 & 60.8 & 70.0 & 41.9 & 54.0 & 54.6 & 52.1 & { 45.1} \\
ResNet-50* & 34.0 & 35.9 & 35.7 & 32.3 & 31.5 & 45.9 & 51.1 & 50.4 & 43.8 & 59.1 & 67.9 & 41.0 & 56.5 & 59.8 & 54.4 & { 46.6} \\
\textbullet~Combined & \textbf{37.2} & \textbf{38.1} & \textbf{39.8} & \textbf{35.9} & \textbf{33.7} & \textbf{50.2} & \textbf{55.3} & \textbf{55.6} & \textbf{50.1} & \textbf{63.8} & \textbf{72.3} & \textbf{45.9} & \textbf{59.6} & \textbf{62.1} & \textbf{58.0} & { \textbf{50.5}} \\ \midrule
Mobile-ViT & 30.9 & 33.5 & 33.5 & 31.8 & 30.5 & 46.2 & 51.5 & 53.3 & 50.1 & 61.1 & 69.6 & 44.8 & 55.0 & 55.2 & 53.0 & { 46.7} \\
ResNet-101* & 37.7 & 39.8 & 38.9 & 36.8 & 36.8 & 49.8 & 54.8 & 54.3 & 49.0 & 61.7 & 69.5 & 47.9 & 60.0 & 62.3 & 57.5 & { 50.5} \\
\textbullet~Combined & \textbf{39.9} & \textbf{42.3} & \textbf{42.0} & \textbf{39.2} & \textbf{38.6} & \textbf{53.4} & \textbf{57.9} & \textbf{59.1} & \textbf{54.8} & \textbf{66.0} & \textbf{73.2} & \textbf{52.1} & \textbf{62.4} & \textbf{63.8} & \textbf{60.1} & { \textbf{53.7}} \\ \midrule
ResNet-50 & 34.6 & 37.6 & 36.7 & 32.9 & 32.7 & 47.9 & 52.5 & 51.4 & 41.1 & 59.8 & 67.4 & 22.6 & 57.9 & 60.8 & 55.7 & 46.1 \\
ResNet-101* & 36.0 & 39.2 & 38.2 & 34.8 & 35.4 & 49.3 & 55.1 & 53.2 & 43.1 & 61.3 & 69.6 & 24.5 & 60.3 & 62.5 & 57.7 & 48.0 \\
\textbullet~Combined & \textbf{38.7} & \textbf{41.9} & \textbf{41.0} & \textbf{36.8} & \textbf{37.0} & \textbf{52.4} & \textbf{57.3} & \textbf{55.8} & \textbf{45.1} & \textbf{63.7} & \textbf{71.3} & \textbf{25.6} & \textbf{62.5} & \textbf{65.0} & \textbf{60.1} & \textbf{50.3} \\ \midrule
Mobile-ViT & 32.6 & 34.7 & 35.8 & 33.1 & 32.2 & 46.9 & 50.4 & 54.0 & 51.4 & 61.4 & 70.7 & 46.5 & 55.0 & 55.0 & 53.4 & 47.5 \\
ViT-Base* & \textbf{55.6} & 55.9 & 56.9 & 57.2 & \textbf{55.7} & 62.6 & 58.8 & 65.5 & 64.5 & 73.0 & 78.2 & \textbf{69.8} & 65.6 & 71.2 & 68.7 & 64.0 \\
\textbullet~Combined & 55.2 & \textbf{56.0} & \textbf{57.1} & \textbf{57.3} & 55.6 & \textbf{63.5} & \textbf{61.1} & \textbf{67.3} & \textbf{65.5} & \textbf{73.7} & \textbf{78.8} & 69.0 & \textbf{67.5} & \textbf{71.3} & \textbf{68.8} & \textbf{64.5} \\ \midrule
Mobile-ViT & 31.6 & 34.8 & 35.3 & 32.1 & 32.0 & 47.2 & 51.9 & 53.4 & 50.8 & 61.6 & 70.3 & 45.7 & 55.1 & 55.6 & 53.1 & 47.4 \\
ViT-Large* & 66.0 & 67.4 & 68.2 & 63.6 & 63.4 & 70.4 & 67.9 & 75.8 & 71.6 & 77.4 & \textbf{83.7} & 75.7 & 71.4 & 77.3 & \textbf{75.6} & 71.7 \\
\textbullet~Combined & \textbf{66.4} & \textbf{67.8} & \textbf{68.6} & \textbf{64.0} & \textbf{63.8} & \textbf{70.4} & \textbf{69.0} & \textbf{76.2} & \textbf{72.4} & \textbf{77.8} & 83.4 & \textbf{76.1} & \textbf{72.8} & \textbf{76.8} & 75.2 & \textbf{72.0} \\ \midrule
ResNet-50 & 42.5 & 44.5 & 43.5 & 41.0 & 40.5 & 52.2 & 54.4 & 55.0 & 49.7 & 61.6 & 68.5 & 51.8 & 59.8 & 62.7 & 57.6 & 52.3 \\
ViT-Large* & 66.3 & 67.7 & 68.7 & 63.8 & 64.1 & 70.9 & 68.5 & 76.0 & 71.7 & 77.5 & 83.8 & 76.1 & 71.6 & 77.8 & 75.9 & 72.0 \\
\textbullet~Combined & \textbf{67.1} & \textbf{68.5} & \textbf{69.3} & \textbf{64.8} & \textbf{64.9} & \textbf{71.9} & \textbf{70.2} & \textbf{76.7} & \textbf{72.3} & \textbf{78.2} & \textbf{83.9} & \textbf{76.4} & \textbf{73.6} & \textbf{78.6} & \textbf{76.7} & \textbf{72.9} \\ \midrule
ViT-Base & 58.2 & 58.7 & 59.2 & 59.9 & 58.9 & 64.4 & 60.5 & 66.4 & 64.5 & 73.7 & 78.6 & 71.5 & 66.4 & 72.6 & 70.1 & 65.6 \\
ViT-Large* & 66.4 & 67.5 & 68.4 & 63.8 & 63.8 & 70.5 & 66.6 & 74.3 & 71.6 & 77.1 & 83.7 & 76.1 & 70.3 & 77.4 & 75.6 & 71.5 \\
\textbullet~Combined & \textbf{67.7} & \textbf{68.8} & \textbf{69.4} & \textbf{65.9} & \textbf{65.5} & \textbf{71.7} & \textbf{67.8} & \textbf{74.8} & \textbf{72.4} & \textbf{78.5} & \textbf{83.8} & \textbf{77.3} & \textbf{72.3} & \textbf{78.3} & \textbf{76.1} & \textbf{72.7} \\ \midrule
ResNet-18 & 29.0 & 30.9 & 30.3 & 25.4 & 26.3 & 39.1 & 44.7 & 43.2 & 32.9 & 52.2 & 60.2 & 16.3 & 50.5 & 53.5 & 49.1 & { 38.9} \\
ResNet50* & 32.4 & 34.7 & 34.4 & 29.2 & 29.6 & 44.8 & 51.1 & 49.6 & 38.3 & 58.7 & 67.6 & 19.3 & 56.7 & 59.8 & 54.3 & { 44.0} \\
\textbullet~Combined & \textbf{34.8} & \textbf{37.1} & \textbf{36.6} & \textbf{31.0} & \textbf{31.5} & \textbf{46.8} & \textbf{52.4} & \textbf{51.3} & \textbf{39.6} & \textbf{60.0} & \textbf{68.0} & \textbf{20.1} & \textbf{58.1} & \textbf{61.0} & \textbf{56.1} & { \textbf{45.6}} \\ \midrule
ResNet-50 & 41.6 & 43.2 & 43.2 & 40.6 & 39.5 & 51.4 & 48.5 & 50.7 & 42.3 & 61.5 & 68.4 & 51.5 & 58.5 & 62.4 & 57.2 & 50.7 \\
ViT-Base & 56.4 & 56.7 & 57.6 & 58.2 & 56.5 & 62.7 & 55.9 & 61.9 & 53.6 & 73.2 & 78.1 & 70.1 & 66.0 & 72.0 & 69.0 & 63.2 \\
\textbullet~Combined & \textbf{58.3} & \textbf{58.8} & \textbf{59.6} & \textbf{59.5} & \textbf{57.9} & \textbf{64.9} & \textbf{58.4} & \textbf{63.9} & \textbf{54.9} & \textbf{74.3} & \textbf{78.8} & \textbf{70.8} & \textbf{68.9} & \textbf{73.6} & \textbf{70.6} & \textbf{64.9} \\ \midrule
ResNet-18 & 30.7 & 33.5 & 30.8 & 27.9 & 27.4 & 40.0 & 45.3 & 43.8 & 34.0 & 52.4 & 60.3 & 20.1 & 50.9 & 53.6 & 49.4 & { 40.0} \\
ResNet-101* & 36.4 & 39.3 & 37.1 & 34.4 & 33.9 & 48.7 & 54.5 & 52.6 & 41.8 & 61.0 & 69.3 & 24.8 & 60.2 & 62.4 & 57.8 & { 47.6} \\
\textbullet~Combined & \textbf{38.7} & \textbf{41.4} & \textbf{39.1} & \textbf{35.9} & \textbf{35.2} & \textbf{50.1} & \textbf{55.6} & \textbf{53.8} & \textbf{42.9} & \textbf{62.1} & \textbf{69.8} & \textbf{25.8} & \textbf{61.2} & \textbf{63.5} & \textbf{58.9} & { \textbf{48.9}} \\ \midrule
ResNet-18 & 34.3 & 36.2 & 34.8 & 32.8 & 32.6 & 41.9 & 43.7 & 42.7 & 37.3 & 53.4 & 60.7 & 42.3 & 51.0 & 54.5 & 50.3 & { 43.2} \\
{ ViT-Base*} & 55.8 & 56.1 & 57.0 & 57.7 & 56.2 & 62.2 & 57.5 & 62.0 & 57.6 & 72.9 & 77.9 & 69.8 & 65.3 & 71.7 & 68.7 & { 63.2} \\
\textbullet~Combined & \textbf{56.8} & \textbf{57.5} & \textbf{58.1} & \textbf{58.7} & \textbf{57.1} & \textbf{63.6} & \textbf{59.4} & \textbf{63.4} & \textbf{58.2} & \textbf{73.6} & \textbf{77.9} & \textbf{70.1} & \textbf{67.3} & \textbf{72.6} & \textbf{69.6} & { \textbf{64.3}} \\ \midrule
ResNet-18 & 34.8 & 36.7 & 34.8 & 33.0 & 33.1 & 42.4 & 46.2 & 45.8 & 41.0 & 53.2 & 60.9 & 42.2 & 51.5 & 54.6 & 50.6 & { 44.1} \\
ViT-Large* & 66.3 & 67.6 & 68.4 & 63.6 & 63.5 & 70.8 & 67.9 & 76.0 & 71.7 & 77.3 & 83.9 & 76.0 & 71.5 & 77.8 & 75.8 & { 71.9} \\
\textbullet~Combined & \textbf{66.8} & \textbf{68.1} & \textbf{68.9} & \textbf{64.2} & \textbf{64.1} & \textbf{71.3} & \textbf{69.0} & \textbf{76.1} & \textbf{72.0} & \textbf{77.8} & \textbf{83.8} & \textbf{75.9} & \textbf{72.8} & \textbf{78.3} & \textbf{76.2} & { \textbf{72.3}} \\ \midrule
ResNet-101 & 42.2 & 43.8 & 43.6 & 41.8 & 41.7 & 50.3 & 52.8 & 53.7 & 49.3 & 59.6 & 65.7 & 49.8 & 57.6 & 59.8 & 56.3 & \multicolumn{1}{l}{51.2} \\
ViT-Base* & 57.2 & 58.2 & 58.5 & 59.2 & 58.3 & 64.2 & 61.6 & 67.7 & 66.7 & 74.0 & 79.1 & 70.6 & 68.1 & 73.1 & 70.7 & \multicolumn{1}{l}{65.8} \\
\textbullet~Combined & \textbf{58.7} & \textbf{60.2} & \textbf{60.1} & \textbf{60.5} & \textbf{59.5} & \textbf{66.1} & \textbf{64.2} & \textbf{69.4} & \textbf{67.5} & \textbf{74.8} & \textbf{79.0} & \textbf{70.9} & \textbf{70.3} & \textbf{74.3} & \textbf{71.7} & \multicolumn{1}{l}{\textbf{67.1}}\\ \midrule

ResNet-101 & 46.1 & 48.1 & 47.4 & 45.2 & 45.5 & 56.1 & 57.9 & 58.8 & 53.2 & 64.1 & 70.3 & 55.0 & 62.7 & 65.1 & 60.6 & { 55.7} \\
ViT-Large* & 66.3 & 67.8 & 68.6 & 63.8 & 64.2 & 71.0 & 68.3 & 76.1 & 71.7 & 77.4 & 83.9 & 76.0 & 71.7 & 77.9 & 76.0 & { 72.1} \\
\textbullet~Combined & \textbf{67.3} & \textbf{68.8} & \textbf{69.3} & \textbf{65.0} & \textbf{65.4} & \textbf{72.2} & \textbf{70.3} & \textbf{76.8} & \textbf{72.4} & \textbf{78.4} & \textbf{84.0} & \textbf{76.4} & \textbf{74.3} & \textbf{78.7} & \textbf{76.8} & { \textbf{73.1}} \\ \midrule

\end{tabular}%
        }
    \end{threeparttable}
    \end{center}
    \vspace{-0.2in}
    \caption{Comparison of accuracy (\%) on ImageNet-C (level 5) with pairwise collaboration of all models. An asterisk (*) denotes the anchor model.}
    \label{allmodelssup}
\end{table*}


%

\section{The Applicability of COCA to More Than Two Models}\label{suppl:sec:three-models}
The general procedure follows a similar structure to the pseudocode outlined in Algorithm 1. Given three models, denoted as M1, M2, and M3, we rank them by size in descending order: M1 $>$ M2 $>$ M3. The co-learning process is conducted in two steps: 1) M2 and M3 are designated as the anchor model and auxiliary model, respectively. COCA is trained using these two models. 2) M1 and the combination of M2 and M3 are designated as the anchor model and auxiliary model, respectively. COCA is then re-trained using this setup to produce the final performance results. The results of this process are summarized in Table~\ref{multimodles}, demonstrating an improvement in average performance across all conditions, highlighting COCA’s effectiveness across three models.  


We further investigate the applicability of COCA by incorporating a third small model to collaborate with a pair of small models. To diversify the model selection, we include PiT~\cite{heo2021rethinking}, a pooling-based Vision Transformer, which differs from both ViT and ResNet. PiT has 10.6M parameters and achieves 31.8\% accuracy (averaged across 15 corruptions in ImageNet-C) with Tent~\cite{wang2020tent}. As shown in Table~\ref{multimodles}, incorporating a third model further enhances the overall performance.

\vfill

    \clearpage

\begin{table*}[t]
    \setlength{\tabcolsep}{3pt} 
    \renewcommand{\arraystretch}{1.2} 
    \begin{center}
    \begin{threeparttable}
        \resizebox{\linewidth}{!}{%
    \begin{tabular}{c|ccc|cccc|cccc|cccc|c}
 \multicolumn{1}{c}{}& \multicolumn{3}{c}{Noise} & \multicolumn{4}{c}{Blur} & \multicolumn{4}{c}{Weather} & \multicolumn{4}{c}{Digital} &  \\ \midrule
Models & Gauss. & Shot & Impul. & Defoc. & Glass & Motion & Zoom & Snow & Frost & Fog & Brit. & Contr. & Elastic & Pixel & JPEG & Avg. \\ \midrule
RN-50 + ViT-B* & 58.3 & 58.8 & 59.6 & 59.5 & 57.9 & 64.9 & 58.4 & 63.9 & 54.9 & 74.3 & 78.8 & 70.8 & 68.9 & 73.6 & 70.6 & 64.9 \\ \midrule 
\textcolor{blue}{RN-18} + RN-50 + ViT-B* & 58.8 & 59.8 & 60.2 & 59.2 & 59.0 & 66.3 & 65.7 & 69.5 & 61.0 & 75.1 & 78.5 & 69.5 & 71.3 & 74.4 & 71.8 & 66.7 \\ \midrule \midrule
RN-101 + ViT-B* & 58.8 & 59.3 & 59.7 & 59.7 & 58.9 & 65.6 & 59.7 & 63.9 & 55.0 & 74.4 & 79.0 & 71.3 & 69.7 & 74.0 & 71.1 & 65.3 \\ \midrule 
\textcolor{blue}{RN-50} + RN-101 + ViT-B* & 59.7 & 60.9 & 61.0 & 60.0 & 60.1 & 67.8 & 67.0 & 70.6 & 60.0 & 75.6 & 79.2 & 70.4 & 72.5 & 75.0 & 72.5 & 67.5 \\ \midrule
\textcolor{blue}{M-ViT} + RN-101 + ViT-B* & 56.7 & 57.7 & 58.2 & 57.7 & 57.2 & 65.6 & 65.0 & 69.9 & 61.3 & 74.5 & 79.0 & 68.5 & 70.9 & 73.3 & 70.4 & 65.7 \\ \midrule \midrule
ViT-B + ViT-L* & 67.7 & 68.8 & 69.4 & 65.9 & 65.5 & 71.7 & 67.8 & 74.8 & 72.4 & 78.5 & 83.8 & 77.3 & 72.3 & 78.3 & 76.1 & 72.7 \\ \midrule 
\textcolor{blue}{M-ViT} + ViT-B + ViT-L* & 68.7 & 70.1 & 70.7 & 67.3 & 67.5 & 71.2 & 70.0 & 75.5 & 73.0 & 78.8 & 83.0 & 77.7 & 74.8 & 77.8 & 75.5 & 73.4 \\ \midrule
\textcolor{blue}{RN-101} + ViT-B + ViT-L* & 68.8 & 69.8 & 70.4 & 67.2 & 67.8 & 73.4 & 72.0 & 77.3 & 62.3 & 79.9 & 83.8 & 77.2 & 76.8 & 80.0 & 77.6 & 73.6 \\ \midrule \midrule
RN-34 + M-ViT* & 24.2 & 27.5 & 25.3 & 25.7 & 29.7 & 41.8 & 50.9 & 49.7 & 48.3 & 59.0 & 71.1 & 36.6 & 52.7 & 52.0 & 49.3 & 43.0 \\ \midrule 
\textcolor{blue}{RN-18} + RN-34 + M-ViT* & 35.4 & 36.8 & 37.0 & 34.5 & 34.0 & 51.2 & 54.3 & 56.5 & 48.4 & 63.6 & 71.8 & 47.3 & 60.1 & 62.1 & 58.2 & 50.1 \\ \midrule \midrule
M-ViT + RN-18* & 34.9 & 37.3 & 36.7 & 33.8 & 33.0 & 47.6 & 53.3 & 53.7 & 48.7 & 61.7 & 70.4 & 43.9 & 57.6 & 59.2 & 56.0 & 48.5 \\ \midrule 
RN-18 + M-ViT + \textcolor{blue}{PiT*} & 37.4 & 47.1 & 47.5 & 42.8 & 38.5 & 55.3 & 53.4 & 60.5 & 45.6 & 66.8 & 73.5 & 57.1 & 62.6 & 64.7 & 62.4 & 54.5 \\ \midrule
\end{tabular}
        }
    \end{threeparttable}
    \end{center}
    \vspace{-0.2in}
    \caption{Experimental results for COCA on ImageNet-C (severity level 5, \%) across three models are presented. In each group, the newly introduced third model is highlighted in blue. Notably, adding a smaller model generally boosts the performance of the two larger models. Here, for brevity, RN-18, RN-34, RN-50, and RN-101 denote ResNet-18, ResNet-34, ResNet-50, and ResNet-101, respectively; ViT-B and ViT-L denote ViT-Base and ViT-Large, respectively; and M-ViT represents Mobile-ViT.}
    \label{multimodles}
\end{table*}

\begin{figure*}[h]
\end{figure*}
\begin{figure*}[h]
\end{figure*}
\begin{figure*}[h]
\end{figure*}
\begin{figure*}[h]
\end{figure*}
\begin{figure*}[h]
\end{figure*}
\begin{figure*}[h]
\end{figure*}
\begin{figure*}[h]
\end{figure*}
\begin{figure*}[h]
\end{figure*}

\end{document}